\documentclass[acmsmall]{acmart}

\AtBeginDocument{%
  }

\setcopyright{acmlicensed}
\copyrightyear{2024}
\acmYear{2024}
\acmDOI{XXXXXXX.XXXXXXX}

\acmJournal{JACM}
\acmVolume{37}
\acmNumber{4}
\acmArticle{1}
\acmMonth{8}

\usepackage{soul}
\usepackage{booktabs} % For formal tables
\usepackage{graphicx}
\usepackage{subfigure}
\usepackage{amsthm}
\usepackage{amsmath}
\usepackage{multirow}
\usepackage{enumerate}
\usepackage{amsfonts}
\usepackage{color}
\usepackage{url}
\usepackage{amsfonts}
\usepackage{tabularx}
\usepackage{diagbox}
\usepackage{longtable}
\usepackage{rotating}
\usepackage{slashbox}
\usepackage[justification=centering]{caption}
\usepackage[ruled, linesnumbered]{algorithm2e}
\usepackage{bm}
\usepackage{float}
\usepackage{makecell}
\usepackage{soul}
\begin{document}

\title{LLM-Enhanced User-Item Interactions: Leveraging Edge Information for Optimized Recommendations}

\author{Xinyuan Wang}
\email{xwang735@asu.edu}
\affiliation{%
  \institution{Arizona State University}
  \city{Tempe}
  \state{Arizona}
  \country{USA}
}

\author{Liang Wu}
\affiliation{%
  \institution{Coupang}
  \city{Mountain View}
  \country{USA}}
\email{liwu5@coupang.com}

\author{Liangjie Hong}
\affiliation{%
  \institution{Linkedin}
  \country{USA}
}
\email{liahong@linkedin.com}

\author{Hao Liu}
\affiliation{%
 \institution{HKUST (Guangzhou)}
 \city{Guangzhou}
 \state{Guangdong}
 \country{China}}
\email{liuh@ust.hk}

\author{Yanjie Fu}
\affiliation{%
  \institution{Arizona State University}
  \city{Tempe}
  \state{Arizona}
  \country{USA}}
\email{yanjie.fu@asu.edu}

\renewcommand{\shortauthors}{Wang et al.}

\begin{abstract}
Graph recommendation methods, representing a connected interaction perspective,  reformulate user-item interactions as graphs to leverage graph structure and topology to recommend and have proved practical effectiveness at scale. 
Large language models, representing a textual generative perspective, excel at modeling user languages, understanding behavioral contexts, capturing user-item semantic relationships, analyzing textual sentiments, and generating coherent and contextually relevant texts as recommendations. 
However, there is a gap between the connected graph perspective and the text generation perspective as the task formulations are different. A research question arises: how can we effectively integrate the two perspectives for more personalized recsys? 
To fill this gap, we propose to incorporate graph-edge information into LLMs via prompt and attention innovations. 
We reformulate recommendations as a probabilistic generative problem using prompts. 
We develop a framework to incorporate graph edge information from the prompt and attention mechanisms for graph-structured LLM recommendations. 
We develop a new prompt design that brings in both first-order and second-order graph relationships; we devise an improved LLM attention mechanism to embed direct the spatial and connectivity information of edges. 
Our evaluation of real-world datasets demonstrates the framework's ability to understand connectivity information in graph data and to improve the relevance and quality of recommendation results. 
Our code is released at: \url{https://github.com/anord-wang/LLM4REC.git}. 

\end{abstract}  





\maketitle

% \vspace{-0.3cm}
\section{Introduction}
% \vspace{-0.1cm}
%%%%%%BY XINYUAN
%We aim to explore new methods for applying large language models to recommendation systems, especially when dealing with data containing a large amount of edge information.  Our goal is to develop a recommendation system that can effectively utilize this edge information while integrating the powerful text-processing capabilities of LLM.  Through this approach, not only can we improve the relevance and accuracy of recommendations, but we can also provide a richer and more personalized experience.
%%%%%%%%%%%%%%%%%%%%%%
Recommender Systems (RecSys) analyzes multi-source user behavioral data and models complex user-item interactions for accurate and personalized recommendations. In networked data, graphs play a role in RecSys due to their ability to represent user-item relationships and capture complex user-user and item-item connections. Example methods include graph neural networks, random walks, matrix factorization, and PageRank ~\cite{brin1998anatomy}. In textual data, Large Language Models (LLMs) have demonstrated amazing abilities to model user languages, understand behavioral contexts, capture user-item semantic relationships, analyze textual sentiments, and generate coherent and contextually relevant texts as recommendations. In this paper, we aim to integrate graphs (a connected interaction aspect) and LLMs (a textual generative aspect) to improve the relevance,  accuracy, and personalization of RecSys. 

%%%%%%%%%%%%%%%%%%%%%
%In recent years, significant progress has been made in the development of LLMs such as BERT and GPT. These models demonstrate excellent performance in natural language understanding and generation, showing enormous potential in applications in multiple fields ~\cite{han2021pre}~\cite{peng2023study}. With the advancement of technology, LLM's ability to handle complex text data continues to enhance, providing new solutions for various tasks and opening a new chapter in intelligent system research~\cite{Birhane2023}. In this context, combining LLM with recommendation systems has become a cutting-edge and revolutionary research field ~\cite{lin2023can}. Traditional recommendation systems focus on analyzing users' behavior data, while when combined with LLM, recommendation systems can understand explicit feedback from users and mine deeper into their implicit needs and preferences ~\cite{ren2023representation}. This combination provides new possibilities for improving the accuracy and satisfaction of recommendation systems.
%%%%%%%%%%%%%%%%%%%%%%

In prior literature, graph-based methods, such as random-walk ~\cite{page1999pagerank}, graph neural networks ~\cite{kipf2016semi}, graph factorization ~\cite{ahmed2013distributed}, and graph contrastive learning ~\cite{you2020graph} have reformulated RecSys as a task of link prediction or graph classification and have demonstrated the effectiveness of such connected graph perspective. 
Recently, LLMs have seen remarkable advancements. The Transformer Architecture and Attention Mechanisms~\cite{vaswani2017attention} was introduced in 2017, followed by large-scale pre-trained language models like BERT ~\cite{devlin2018bert}. In 2018,  OpenAI introduced the GPT (Generative Pre-trained Transformer) model ~\cite{radford2018improving}, which is one of the first large-scale language models based on the Transformer Encoder Architecture. Subsequent improved versions, such as GPT-2 (2019) ~\cite{radford2019language},  GPT-3 (2020) ~\cite{brown2020language}, GPT-4 (2023) ~\cite{achiam2023gpt}, and GPT-4o (2024), increased model size, leading to improved performance and smarter capabilities like few-shot learning and open-ended text generation.
Researchers found that the LLMs, with their ability to capture rich contextual representations and handle long-range dependencies, showed potential in understanding user preferences and item characterizations. 
Geng et al. proposed the concept of Recommendation as Language Processing in 2022 and leveraged pre-trained LLMs like T5~\cite{raffel2020exploring} to perform sequential recommendation, rating prediction, explanation generation, and direct recommendation. ~\cite{geng2022recommendation}
Thereafter, there are many studies that integrate LLM into RecSys by improving prompt engineering, fairness, and debiasing, efficient model architectures, explainability, ethical issues under RecSys ~\cite{zhao2022resetbert4rec, sun2019bert4rec, hou2022towards, liu2023chatgpt, dai2023uncovering}. 

%%%%%%%%%%%%%%%%%%%
%However, a major challenge in RecSys is that although data contain a large amount of edge information (such as the relationship between users and items), this information is not fully utilized in LLMs, especially in its key attention mechanism~\cite{wang2023enhancing}. Even though people have used this edge information to construct a variety of prompts ~\cite{wu2023survey}, existing methods of building recommendation systems using LLMs cannot consider edge information structurally. This raises a research question: how to effectively integrate graph information within the framework of LLMs to improve the performance of recommendation systems?
%%%%%%%%%%%%%%%%%%
However, there is a clear gap between LLMs-based RecSys and graph-based RecSys. In the LLMs setting, RecSys is seen as an auto-regressive textual generation task, while in the graph setting, RecSys is seen as a link prediction task. The two perspectives are fundamentally different. A major challenge is how we can effectively integrate the connected graph perspective (edge information) into LLMs for effective RecSys.

\noindent\textbf{Our Insights: Incorporating Graph Edge Information into Large Language RecSys via Prompt and Attention Innovations.}
To fill the gap between LLMs-based RecSys and graph based RecSys, we renovate prompt design and attention mechanism.
Firstly, we develop a prompt design to take into account not just user profiles and item descriptions, but also user-item edge interactions that particularly include both direct relationships between users and items and second-order relationships between items, a complex association not directly present in traditional recommendation data.
Such graph edge-enhanced prompt design equips LLMs with better abilities of contextual understanding of user-item relationships. 
Secondly, we introduce a graph structure knowledge attention mechanism into the pre-trained transformer model. The attention mechanism leverages not just 
the relationships between nodes by modeling their connectivity
(direct relationship) but also spatial information (indirect relationship)
in the graph.
%By incorporating graph and edge inforamtion into prompt design and model attentions, we advance the contextual learning ability of LLMs for accurate recommendation generation. 

\noindent\textbf{Solution Summary.}
We regard user and item nodes as target recommendation tokens and reformulate recommendations as a probabilistic generative problem using prompts.
Specifically, we first construct a user-item interaction graph and develop a graph structure-guided attentive LLM backbone with a new neural attentive decoder to model these connections.
The training is achieved by creating text prompts for all users and items, including interaction events and crowd contextual prompts, using these prompts to pre-train the backbone by maximizing text generative likelihood to learn contextual knowledge relevant to recommendations.
Besides, we convert a user’s interaction history into past-tense texts and combine them with future-tense triggers to fine-tune the LLM backbone, by minimizing recommendation errors.
Finally, we leverage the fine-tuned graph attentive LLM backbone to make recommendations based on the personalized rating or purchase history and predictive triggers of test users.

%%%%%%%%%%%%%%%%%%%
\iffalse
\begin{itemize}
  \item An innovative way of combining LLMs with recommender systems: By creatively integrating LLMs' deep text understanding ability with users' behavior analysis of recommender systems, our method can more comprehensively understand users' and items' information. 
  Our method effectively utilizes the language processing capabilities of LLMs, bringing new dimensions and depth to traditional recommendation logic, thereby ensuring recommendation quality while also increasing diversity and innovation.
  \item A new prompt strategy: We introduce a new prompt mechanism that can transform the relationship between users and items, as well as the background information of items, into natural language form.  In addition, by constructing second-order relationships between items, we can uncover deeper correlations between items, thereby providing more comprehensive and detailed recommendations.
  \item A novel fusion method for edge information: We propose a new approach that directly embeds the edge information of graph data into the attention mechanism of LLMs. 
  This method effectively utilizes the connection information in the graph structure, enhancing the model's ability to handle complex user-item interactions. 
\end{itemize}
\fi
%%%%%%%%%%%%%%%%%%%%%%

\noindent\textbf{Our Contributions.} 
\ul{\textit{1) Framework.}} We tackle a graph-structured LLM RecSys task and develop a framework to incorporate graph edge information from the prompt and the attention mechanism.  
\ul{\textit{2) Computing.}} We develop a new prompt design that brings in both first-order and second-order graph relationships; we devise an improved LLM attention mechanism to embed direct the spatial and connectivity information of edges. 
\ul{\textit{3) Experiments.}} We present extensive experiments on a series of recommendation datasets to demonstrate the performance of recommendation tasks and the effectiveness of graph-structured prompt and attention mechanisms.

% \vspace{-0.3cm}
\section{Preliminaries}

\subsection{Symbol Definition}

\textbf{Table~\ref{tab:SymbolsTable}} summarizes the key symbols used throughout this paper, covering user-item interactions, textual representations, model parameters, and graph structures. It provides a reference for understanding the mathematical formulations and concepts in our proposed approach.

\begin{table*}[htbp]
\centering
\caption{List of Symbols Used in This Paper.}
\label{tab:SymbolsTable}
\resizebox{0.8\textwidth}{!}{
\begin{tabular}{ll}
\toprule
\textbf{Symbol} & \textbf{Description} \\
\midrule
$I$ & The number of users in the dataset. \\
$J$ & The number of items in the dataset. \\
$X_{ij}$ & The binary interaction between user $i$ and item $j$. \\
$T_{i}$ & The descriptions of the user $i$. \\
$T_{j}$ & The descriptions of the item $j$. \\
$T_{ij}$ & The joint texts of the user $i$ and item $j$. \\
$T$ & The textual descriptions. \\
$N$ & The number of sequences in $T$. \\
$T_{nk}$ & The $k\textit{-}th$ token in the $n\textit{-}th$ sequence. \\
$Q$ & The query vector in the Transformer structure. \\
$K$ & The key vector in the Transformer structure. \\
$V$ & The value vector in the Transformer structure. \\
$\sqrt{d_k}$ & The dimensionality (size) of the key vector to enforce a normalization effect. \\
$R$ & The relationship encoding extracted from graph knowledge. \\
$R^{\text{conn}}$ & The direct connection relationships between nodes. \\
$R^{\text{path}}$ & The normalized shortest path score between nodes. \\
$P$ & The shortest path matrix. \\
$P_{ij}$ & The minimum path length among all possible paths from node $i$ to node $j$. \\
$\delta$ & The normalization factor between 0 and 1. \\
$\mathcal{L}_{\text{pre-train}}$ & The objective function for maximizing text generation likelihood during pre-training. \\
$t_{i}$ & The tokens in the next token prediction task. \\
$\mathcal{L}_{\text{fine-tune}}$ & The objective function for optimizing recommendation accuracy during fine-tuning. \\
$\mathcal{S}_i$ & The list of recommended items $\mathcal{S}_i$ to user $i$. \\
$\text{Pr}_{i}$ & The personalized predictive prompt for user $i$ in fine-tuning. \\
$\Theta$ & The set of LLM parameters. \\
\bottomrule
\end{tabular}
}
\end{table*}

\subsection{Important Definitions}
% \vspace{-0.1cm}
\textbf{Generative Large Language Models.}
Generative Large Language Models (LLMs) are a type of model based on transformer encoders that generate natural language texts~\cite{radford2019language}. LLMs are trained on massive text corpora and can capture broad contextual relationships between words. LLMs generate a series of words \( (w_1, w_2, \ldots, w_n) \) by modeling the joint probability of word sequences \( P(w_1, w_2, \ldots, w_n) \).

\noindent\textbf{Token and Embedding.}
In NLP, tokens are the smallest units for LLMs, such as words, sub-words, or characters~\cite{webster1992tokenization}. Embedding is a dense vector representation of a token in a continuous space that encodes language attributes and semantic information~\cite{pilehvar2020embeddings}. In recommendations, we see users and items as unique tokens. 

\noindent\textbf{Prompt.}
Prompts are designed to guide generative LLMs to generate specific responses. They serve as guides for the model to generate text tokens in specific contexts or styles~\cite{liu2023gpt}.

% \vspace{-0.3cm}
\subsection{Problem Statement}
% \vspace{-0.1cm}
Consider the existence of $I$ users and $J$ items, let $X_{ij}$ be the binary interaction (e.g., purchase) matrix between user $i$ and item $j$.  
Besides, we collect user descriptions, item descriptions (e.g., prices, brand, category, title), user reviews for items, and explanations of user purchase reasons.
We denote $T_{i}$ as the descriptions of the user $i$,  $T_{j}$ as the descriptions of the item $j$, $T_{ij}$ as the joint texts of the user $i$ and item $j$, such as user reviews and purchase reasons for items.
We unify all textual descriptions into $T$ that includes $N$ sequences, $k$ indexes the tokens in each sequence, and $T_{nk}$ is the $k\textit{-}th$ token in the $n\textit{-}th$ sequence. 
Our goal is to leverage  LLMs and graphs to develop a generative recommender system that takes a prompt, including a user ID and a user's historical interaction records with items, and generates product recommendations to the user.

% \vspace{-0.3cm}
\section{Leveraging LLM and Graphs for Recommender Systems}
% \vspace{-0.1cm}

\begin{figure*}[htbp] 
\centering
\includegraphics[width=\linewidth]{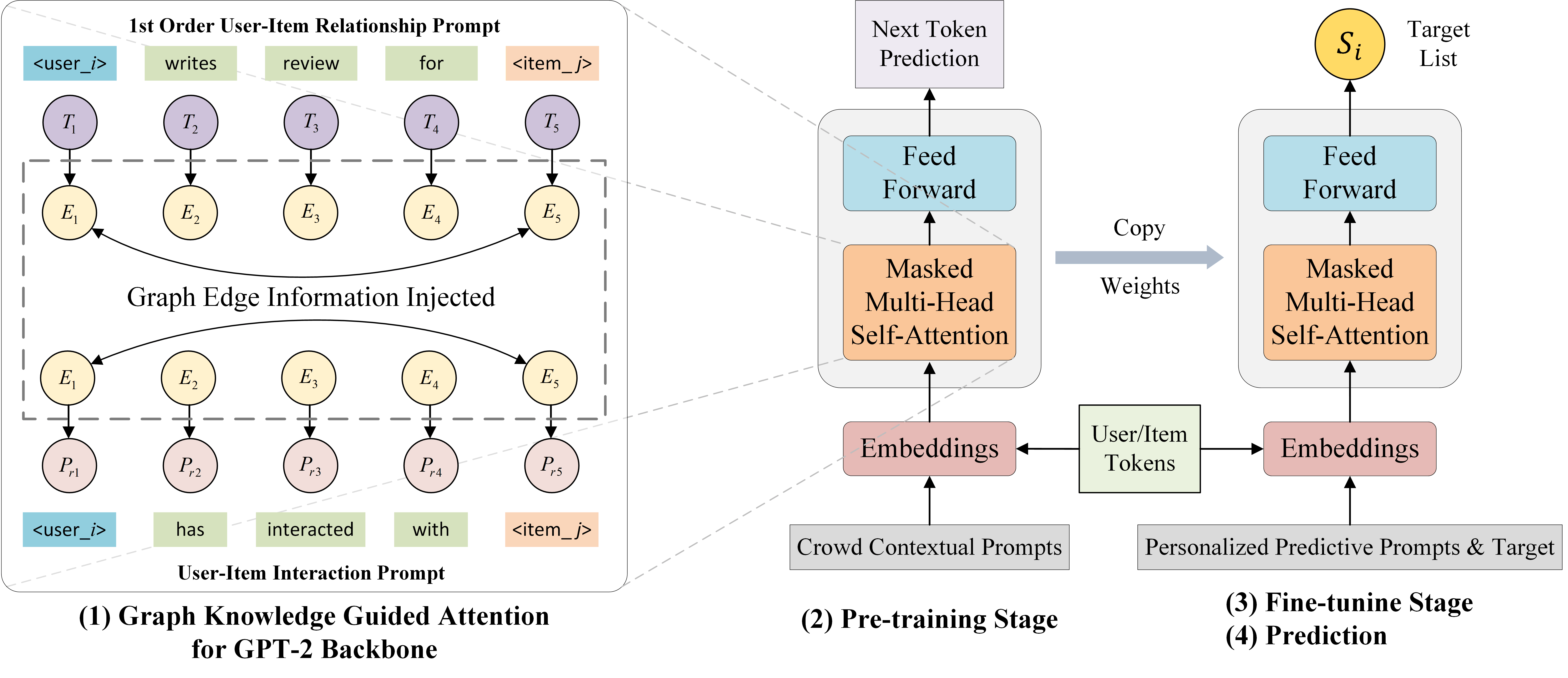} 
% \vspace{-0.6cm}
\captionsetup{justification=raggedright,singlelinecheck=false}
\caption{Overview of the proposed graph-attentive LLM-based recommender system. The framework consists of four key steps: (1) Developing a graph-attentive LLM backbone, which integrates multi-source textual information and user-item interaction graphs to enhance representation learning; (2) Pre-training with crowd contextual prompts, where structured text prompts encode user and item descriptions along with interaction facts to establish contextual knowledge; (3) Fine-tuning with personalized predictive prompts, converting past user interactions into structured prompts with future prediction triggers to optimize recommendation accuracy; and (4) Generative recommendation, where the fine-tuned LLM backbone predicts items based on user history and interaction patterns.}
\label{fig:structure}
% \vspace{-0.5cm}
\end{figure*}

\subsection{Framework Overview}
Given user descriptions, item descriptions, user textual reviews for items, and the user-item interaction (e.g., purchase, rating) graph, we aim to leverage and connect generative LLM, textual generation, and user-item interaction graphs to advance recommender systems~\cite{wu2023survey}.  
\textbf{Fig.~\ref{fig:structure}} shows the four major steps for building our recommender system: 
1) developing graph attentive LLM recommendation backbone; 
2) pre-training the backbone with crowd contextual prompts; 
3) fine-tuning of the backbone with personalized predictive prompts; 
4) generative recommendation. 

In Step 1, considering the existence of multi-source textual information, including user descriptions, item descriptions, and user reviews for items, we propose to leverage LLM to learn the generation of these texts to model the representations of user preferences and item functionalities.  
Aside from texts, the user-item interaction graph can provide two types of signals:  i) user nodes and item nodes as tokens and ii) graph structures about the direct (first-order) connectivity among users and items and the indirect (second-order) connectivity among items, users, or user-item pairs. 
To leverage the user and item tokens, we add the user and item tokens to enrich the texts; to leverage the graph structures, we develop a graph knowledge-guided attentive LLM backbone, particularly with a new neural attentive decoder structure, to model the first-order and second-order connectivity graph knowledge. 
Our graph-attentive LLM backbone reformulates recommendations into a probabilistic generative problem in response to prompts.

Step 2 is to pretrain the graph-attentive LLM backbone. 
Specifically, we first construct the text prompts of all users and items, including the texts of user descriptions, item descriptions, and user reviews for items,  the fact or event checking texts of whether a user interacts with (e.g., rates or purchases) items, as crowd contextual prompts. 
We utilize the crowd contextual prompts to pre-train the backbone by maximizing text generative likelihood. So,  LLM can gain contextual knowledge of a world of recommendation.

Step 3 is to develop personalized prompts of a given user to incentivize the pre-trained backbone to make recommendations. 
Specifically, given a user, we convert the user's interaction history (e.g., ratings, purchases)  with all items into past-tense texts, combined with a future-tense trigger (e.g., user $i$ will purchase $\underset{---}{\text{?}}$), to train the LLM backbone to predict recommendations. The optimization goal is to minimize recommendation errors, not textual generation likelihood.

Finally,  given a test user with the corresponding personalized rating or purchase history and a predictive trigger, Step 4 leverages the fine-tuned graph attentive LLM backbone to recommend items to the test user.

% \vspace{-0.2cm}
\subsection{Graph-Structured Attentive LLM-Based Generative Recommendation Backbone}
% \vspace{-0.1cm}

\subsubsection{GPT2 as LLM Base Model}
\label{backbone}
Our base model is the GPT-2~\cite{radford2019language}. The original GPT-2 utilizes the Transformer architecture, pre-trained on vast text datasets to predict subsequent words in sequences. A multi-layer structure containing attention heads scales up to billions of parameters for enhanced pattern recognition. The model supports conditional text generation and offers various sampling strategies for generating text. 
In GPT-2, the attention mechanism is to weigh the importance of different words in a sentence. It operates by computing attention scores for each word in the input sequence based on their relevance to each other. 
These scores determine how much attention the model should pay to each word when generating the next word in the sequence. By attending to relevant parts of the input text, GPT-2 can capture long-range dependencies and generate coherent and contextually relevant output. 

% \vspace{-0.2cm}
\subsubsection{Integrating Two Structure Knowledge for Graph Attentive LLM}
\label{subsec:InjectedAttention}
When performing generative recommendations, we obtain recommendation results in the form of text generation to connect items to users. 
As a result, users and items are usually regarded as tokens in a text sequence for pre-training. 
In the real world, users interact (e.g., rate, purchase) with items. Such interactions can be modeled as a graph where users or items are seen as nodes, and user-user, item-item, and user-item connectivity is seen as edge weights, representing a kind of graph-structured information propagation-driven edge knowledge.
In other words, user and item tokens are not simply independent entities in a sequence. The LLM should not just learn user and item embeddings by paying attention to their mutual relevance in a sequence. It is critical to leverage the graph-structured edge knowledge to improve LLM for recommendations. 

Firstly, we propose a graph structure knowledge-attentive LLM method to integrate graph knowledge into recommender systems.
Specifically, we incorporate the edge connectivity between users and items into attention weight calculation~\cite{gao2020deep}. Inspired by the Graphormer ~\cite{ying2106transformers}, we leverage Graph Neural Networks (GNNs) to describe the relationships between nodes by modeling their connectivity (direct relationship) and spatial information (indirect relationship) in the graph.
Formally, the edge information, denoted by the $R$-term, is  used to calculate the attention weights in the graph-structured attention mechanism, which is given by: 
\begin{equation}
\text{Attention}(Q,K,V)=\text{SoftMax}\left(\frac{QK^T}{\sqrt{d_k}}+R\right)V,
\label{eq:att}
\end{equation}
where $Q$, $K$, and $V$ are queries, keys, and values respectively, while $R$ represents the relationship encoding extracted from graph knowledge. The $\sqrt{d_k}$ is the dimensionality (size) of the key vector to enforce a normalization effect.
The structural scores $R$ are added to the original attention scores because there should be additional scores between connected nodes. It means that connected nodes should pay more attention to each other. This is a way to reflect graph edges in the attention mechanism.

Secondly, we identify two kinds of important graph structural knowledge: 1) the direct (first-order) connectivity and 2) the indirect (high-order path) connectivity, among users and items. 
Correspondingly, the graph-structured relational attention $R$ term  is composed of two distinct parts of the graph topology: 
\begin{equation}
R = R^{\text{conn}} + R^{\text{path}}. 
\end{equation}
The first part, $R^{\text{conn}}$, represents the direct connection relationships between nodes. Typically, we denote $R^{\text{conn}}$ as  a binary adjacency matrix, which is given by:
\begin{equation}
R_{ij}^{\text{conn}} = \begin{cases}
1, & \text{if there is a direct connection between node } i \text{ and } j \\
0, & \text{otherwise}
\end{cases}.
\end{equation}
In this matrix,  1 indicates there is an edge between two nodes, and 0 indicates there is no edge between two nodes. 
The second part, $R^{\text{path}}$, represents a normalized shortest path score between nodes, which is computed based on the entire graph. The shortest path information is essential because it reflects the indirect relationships of node pairs and the degrees of separation or distance between nodes, which can be highly informative for understanding complex graph structures. 
The normalized shortest path score is calculated using the shortest path matrix $P$, where each element $P_{ij}$ is defined as the minimum path length among all possible paths from node $i$ to node $j$. The $P_{ij}$ is given by:
$
P_{ij} = \text{min}\{ \text{path length} | \text{all paths from node } i \text{ to node } j \}.
$
Later, we introduce a damping factor $\delta$ to adjust the influence of distant nodes. This is achieved by inverting and normalizing the path lengths in the matrix. The modified shortest path matrix $R^{\text{path}}$ is defined as:
\begin{equation}
\label{eq:r_path}
R_{ij}^{\text{path}} = 1 - \frac{\delta^{P{ij}}}{\max(P)}.
\end{equation}
In this formulation, $\delta$ is a value between 0 and 1, and $\max(P)$ represents the maximum path length in the shortest path matrix $P$.
The normalization step ensures that $R_{ij}^{\text{path}}$ remains within the range of 0 to 1. $R_{ij}^{\text{path}}$ captures the proximity between nodes in the graph. Shorter paths (indicating closer connections) result in higher values. 
Integrating the direct connections and indirect relationships between nodes into a unified representation $R$, the attention mechanism is empowered to model the inherent characteristics of individual nodes and their relative positions and interconnections within the overall graph. 

% \vspace{-0.3cm}
\subsection{Pre-training Graph Attentive LLM with Crowd Contextual Prompts}
\label{subsec:Prompt}
% \vspace{-0.1cm}

Pre-training is to initially train our graph-attentive GPT-2 model on a larger corpus of recommendation text data, including data collection, tokenization, model architecture, pre-training objective, and optimization procedure. 

\subsubsection{Data Collection.}
We first collect a large textual corpus from diverse sources: user descriptions, item descriptions, user reviews for items, and historical events that users interact (e.g., rate or purchase) with items, to ensure that the graph-attentive LLM model learns robust representations of languages. 

\subsubsection{The Structure of Crow Contextual Prompts}
\begin{figure}[htbp]
% \vspace{-0.3cm}
\centering
\includegraphics[width=0.7\textwidth]{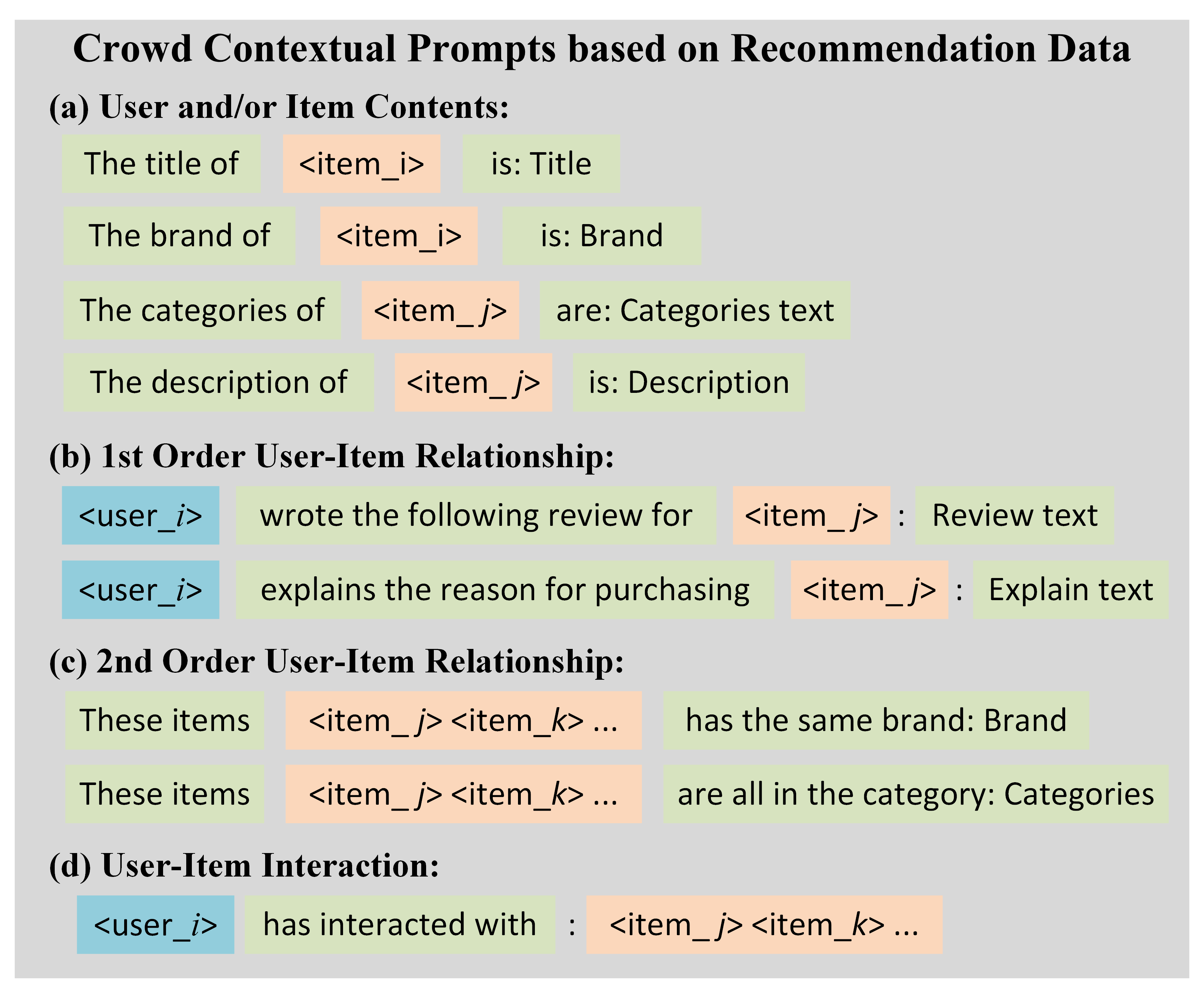}
% \vspace{-0.6cm}
% \caption{Crowd Contextual Prompts based on Recommendation Data.}
\captionsetup{justification=raggedright,singlelinecheck=false}
\caption{The structure of crowd contextual prompts used for pre-training. The prompts include: (1) \textbf{User/Item Tokens} to encode semantic information about users and items; (2) \textbf{Item Contents}, describing item attributes such as title, brand, and categories; (3) \textbf{1st Order User-Item Relationships}, capturing user interactions through reviews and purchase explanations; (4) \textbf{2nd Order User-Item Relationships}, representing item-level associations; and (5) \textbf{User-Item Interaction Events}, incorporating purchase histories as prompts.}

\label{fig:mutilPrompt}
% \vspace{-0.5cm}
\end{figure}

\textbf{Fig.~\ref{fig:mutilPrompt}} shows that we define the unique structure of our crow contextual prompts for pre-training an LLM recommender. 

\noindent\underline{User/Item Tokens.} Our prompts include two unique tokens: user tokens and item tokens. User and item tokens are the targeted entities in a recommender system. In other words, the representation learning and generative recommendation tasks are centered around users and items. As a result, we add userID and itemID tokens into the corpus vocabulary to reflect user and item information.
We expect the embedding of the user ID and item ID to be able to embed rich semantic information about users, such as user profiles, demographics, reviews, and preferences, information propagated from items, and rich semantic information about items, such as item descriptions, item functionalities, item reviews, information propagated from users. 

To keep these special tokens, we revise the original tokenizer to prevent it from decomposing them into smaller ones. If they were broken into smaller pieces, they would have had difficulty precisely representing rich semantic information for users and items.

\begin{itemize}

\item \noindent\underline{User and/or Item Contents.} 
Aside from user IDs and item IDs, we use the content (attributes) information of users and items to develop content-related prompts, such as, the title of an item is <TITLE>, the brand of an item is <BRAND>, and the product categories of an item are <CATEGORIES>. The description of an item is <DESCRIPTIONS>. In our experiments, user contents are removed due to privacy concerns. We only have item contents. 

\item \noindent\underline{1st Order User-Item Relationship.} 
We utilize the historical review comments of users for items to develop first-order user-item relationship prompts, such as, a user wrote the following review for an item: <review texts>, a user explained the reason for buying an item: <explanations>. 

\item \noindent\underline{2nd Order User-Item Relationship.}  
We leverage the second-order information as a type of prompt, such as a few items <ITEM LIST> share the same brand: <BRAND>. 

\item \noindent\underline{User-Item Interaction (e.g., Purchase) Events .} 
Finally, we incorporate the purchase events as prompts, such as a user has interacted with (purchased) <ITEM LIST>. 
\end{itemize}
In this way, we aim to augment the prompt texts and enrich the contextual environment that simulates the preference, characteristics, functionality, categorization,  opinion, and first-order and second-order social or network dimensions of a real recommender system in a language modality. These crowd contextual prompts are used as training data to pre-train our graph-attentive LLM. 

\subsubsection{The Optimization Objective and Procedures of Pre-training}
\label{subsubsec:pretrain}
Given the prepared crowd contextual prompts as pre-training data, we will train the graph attentive GPT-2 to predict the next token in the textual sequence. The optimization objective is to maximize the token generation likelihood of a textual sequence. The likelihood function is given by:
\begin{equation}
\mathcal{L}_{\text{pre-train}} = -\sum_{i} \log P(t_{i+1} | t_{i}, t_{i-1}, \ldots, t_{1}; \Theta),
\label{eq:pretrain}
\end{equation}
where $t_{i}, t_{i-1}, \ldots, t_{1}$ is the first $i$ tokens, and $\Theta$ indicates the weights of the LLM, the objective is to predict the $i+1$ token $t_{i+1}$.
By solving the optimization objective, the graph-attentive LLM can combine knowledge from different information sources to learn more comprehensive portraits of users and items. 
By integrating graph-structured attention, the graph attentive LLM can model the first-order and second-order edge connectivity among users and items.

% \vspace{-0.2cm}
\subsection{Fine-tuning Graph Attentive LLM with Personalized Predictive Prompts}
% \vspace{-0.1cm}
After the pre-training step, the graph-attentive LLM learns the contextual knowledge of users, items, and relationships of a recommender system's world. However, the optimization objective in the pre-training step focuses on maximizing textual generation accuracy in a language sequence instead of recommending personalized items. Therefore, in the fine-tuning step, we develop (1) personalized predictive prompts and (2) recommendation loss functions of fine-tuning to incentivize the graph-attentive LLM to shift model focuses from text generation to generating accurate personalized item recommendations.  

\subsubsection{Personalized Predictive Prompts}
\begin{figure}[htbp]
% \vspace{-0.4cm}
\centering
\includegraphics[width=0.7\textwidth]{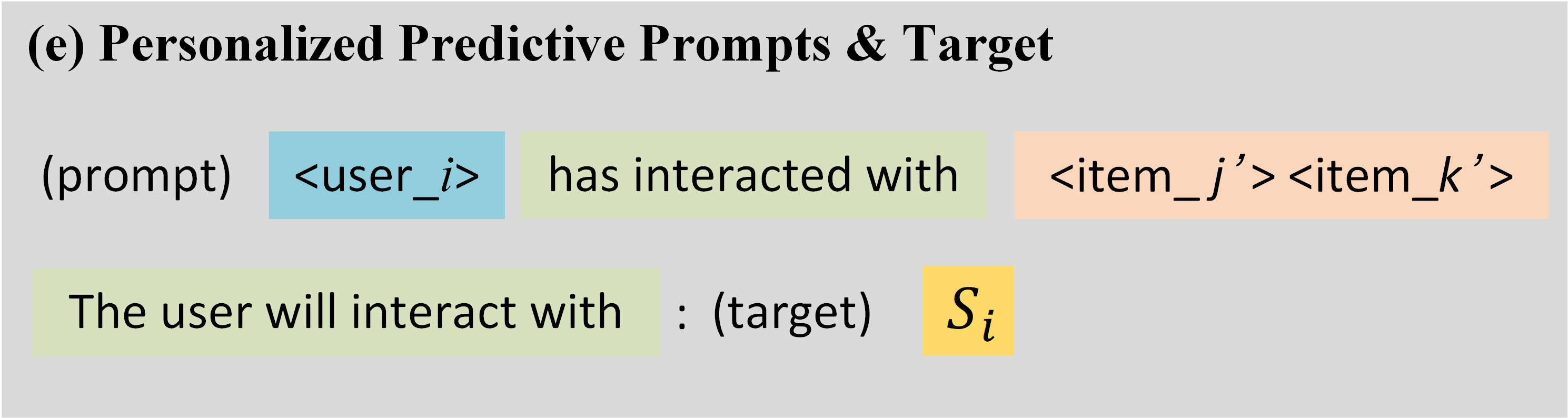}
% \vspace{-0.5cm}
% \caption{A Personalized Predictive Prompt.}
\captionsetup{justification=raggedright,singlelinecheck=false}
\caption{The illustration of the Personalized Predictive Prompt used for fine-tuning. The prompt converts a user’s past interactions into structured past-tense texts, followed by a future-tense trigger. This design encourages the LLM to infer user preferences and generate personalized item recommendations.}

\label{fig:PredictPrompt}
% \vspace{-0.5cm}
\end{figure}

When fine-tuning the pre-trained graph attentive LLM, we introduce the personalized predictive Prompt method.  Our idea is to use the historical purchase events of users for items as prompts to guide the LLM to learn user preferences for items. \textbf{Fig.~\ref{fig:PredictPrompt}} shows that we convert a user’s interaction (e.g., rating, purchases) history with all items into past tense texts, combined with a future tense trigger (e.g., user $i$ will purchase $?$), to motivate the graph attentive LLM to generate item recommendations for a user.

\subsubsection{The Optimization Objective and Training Procedure of Fine-tuning}
\label{subsubsec:finetune}
Different from learning the world knowledge of a recommender system, the fine-tuning stage is to adapt the pre-trained LLM to personalized item recommendations.   During the fine-tuning process, we will integrate personalized predictive prompts into LLM so that the model can model the historical item purchase event of a user to generate a list of items $\mathcal{S}_i$ to the specific user, based on the prompt. The generative recommended items are then compared with the actual purchase records of the specific user.  The resulting loss function derived from this comparison is utilized as the optimization objective of fine-tuning. In particular, we define the generative probability that measures whether LLM recommendations are statistically close to historical user purchase records, which is given by \textbf{Equation~\eqref{eq:finetune}}:
\begin{equation}
\mathcal{L}_{\text{fine-tune}} = -\sum_{i} \log P(\mathcal{S}_i | \text{Pr}_{i}, \Theta),
\label{eq:finetune}
\end{equation}
where $\text{Pr}_{i}$ is the prompt for fine-tuning stage, and $\mathcal{S}_i$ is the final recommendation list, $\Theta$ denotes the LLM weights.
In summary, fine-tuning optimizes the recommendation loss. 

% \vspace{-0.3cm}
% \subsection{Using Fine-tuned Graph Attentive LLM for Item Recommendations.}
% \vspace{-0.1cm}
% After pre-training and fine-tuning, given a testing user $i$, we convert the user's purchase records into a personalized predictive prompt as the input of the graph attentive LLM. The construction of prompts is the same as the fine-tuning stage. 
% Finally, by comparing the probability scores of each item, the top $M$ items with the highest scores are selected as recommendations:
% \begin{equation}
% R_i = \underset{j}{\mathrm{argmax}} \, \left(\text{SoftMax}\left(\mathbf{LLM}(X_{ij}; \Theta)\right)\right),
% \label{eq:predict}
% \end{equation}
% where $X_{ij}$ is the user-item interaction prompt of user $i$ and item $j$. $\Theta$ denotes the weights for the $\mathbf{LLM}$ backbone.

% \vspace{-0.2cm}
\subsection{Graph Attentive LLM for Item Recommendations in Deployment}
% \vspace{-0.1cm}

After pre-training and fine-tuning, given a testing user $i$, we convert the user's interaction records with items into a personalized predictive prompt as the input of the graph-attentive LLM. Therefore, the classic recommendation engine in production can be viewed as a wrapper, where the request is reconstructed as a prompt -the same as the prompt in the fine-tuning stage- and the graph-attentive LLM will provide a recommendation score for each user-item pair.

To reduce the serving latency in production and alleviate the peak load pressure, the proposed graph attentive LLM model can be deployed onto both offline and online GPU clusters. The offline pipeline focuses on the batch processing and calculates the relevance between a user and the candidate items to recommend, shown as $\mathbf{LLM}(\text{Pr}_{i}; \Theta)$, where $\text{Pr}_{i}$ is the user-item interaction prompt of user $i$, and $\Theta$ denotes the weights for the $\mathbf{LLM}$ backbone. The batch processing can be applied directly in offline recommendation scenarios such as promotional emails and notifications. The results can also be used for warming up the online cache to minimize redundant computations. 

In a typical online recommendation scenario, the prompt containing the user, item, and interaction information is sent to the graph attentive LLM model, and the top items with the highest scores are selected as recommendations by comparing the probability scores against each other,
\begin{equation}
\mathcal{S}_i = \underset{j}{\mathrm{argmax}} \,\left(\mathbf{LLM}(\text{Pr}_{i}; \Theta)\right).
\label{eq:predict}
\end{equation}

In this case, the proposed method can easily be integrated into the most common recommender systems in the industry. The additional pressure caused by GPU serving can be handled by offline (batch) pre-computation and online caching warm-up.

\section{Experimental Results}
We conduct empirical experiments to answer the following questions: (1) Can our method generate more accurate recommendations? (2) What are the contributions of different technical components? (3) What are the contributions of second-order relationships and item background information? (4) What are the impacts of different attention mechanisms? (5) parameter sensitivity and robustness. 

\subsection{Experimental Setup}

\subsubsection{Data Description}
We used seven public recommendation datasets: Amazon (AM)-Beauty dataset, AM-Toys dataset, AM-Sports, AM-Luxury, AM-Scientific, and AM-Instruments dataset~\cite{mcauley2016addressing}. 
% These datasets contain users' reviews for items. We build the data we need through pre-processing. 
We binarized the user-item interaction matrix by scores. If the score is greater than 3, there is a connection between a user and an item. 
%Meanwhile, we clean the data and retain the information including interaction, reviews, explanations, and item descriptions. 
For each user in the dataset, we randomly select 80\% of interactions for training, 10\% for validation, and 10\% for testing, with at least one sample selected in both the validation and test sets. According to the prompt construction method in Section~\ref{subsec:Prompt}, we constructed the data for pre-training. \textbf{Table~\ref{tab:Dataset}} shows The main dataset statistics.

\begin{table}[htbp]
% \vspace{-0.3cm}
  \begin{center}
    \caption{Dataset Statistics}
    % \vspace{-0.3cm}
    \label{tab:Dataset}
        \begin{tabular}{lrrrr}
            \toprule
            Dataset & User & Item & Interaction & Content \\
            \midrule
            \textbf{AM-Beauty} & 10,553 & 6,086 & 94,148 & 165,228 \\
            \textbf{AM-Toys} & 11,268 & 7,309 & 95,420 & 170,551 \\
            \textbf{AM-Sports} & 22,686 & 12,301 & 185,718 & 321,887 \\
            \textbf{AM-Luxury} & 2,382 & 1,047 & 21,911 & 15,834 \\
            \textbf{AM-Scientific} & 6,875 & 3,484 & 50,985 & 43,164 \\
            \textbf{AM-Instruments} & 20,307 & 7,917 & 183,964 & 143,113 \\
            \textbf{AM-Food} & 95,421 & 32,180 & 834,514 & 691,543 \\
            \bottomrule
        \end{tabular}
    \end{center}
\end{table}

% \vspace{-0.1cm}
\subsubsection{Evaluation Metrics}
We used three metrics: Recall@20, Recall@40, and NDCG@100 to evaluate algorithmic effectiveness.
Recall@k~\cite{wang2013theoretical} indicates the proportion of items that users are interested in among the top-$k$ recommended items: 
\begin{equation}
\text{Recall@k} = \frac{ | \text{Relevant items} \cap \text{Recommended items at k} | }{ | \text{Relevant items} | },
\end{equation}
NDCG@k is a position-sensitive indicator that measures the quality of recommendation lists: 
\begin{equation}
\text{NDCG@k} = \frac{ \text{DCG@k} }{ \text{IDCG@k} },
\end{equation}
where, $\text{DCG@k} = \sum_{i=1}^{k} \frac{ 2^{rel_i} - 1 }{ \log_2(i+1) }$ and $\text{IDCG@k} = \sum_{i=1}^{|REL|} \frac{ 2^{rel_i} - 1 }{ \log_2(i+1) }$

% \vspace{-0.1cm}
\subsubsection{Baseline Algorithms}
We compared our method with various approaches, including ID-based and Attention-based methods: 
\begin{itemize}
  \item Multi-VAE~\cite{liang2018variational} is an ID-based collaborative filtering method that completes recommendation tasks by using a polynomial likelihood variational auto-encoder to reconstruct ratings.
  \item MD-CVAE~\cite{zhu2022mutually} extends Multi-VAE by introducing dual feature VAE on text features to regularize rating reconstruction.
  \item BERT4REC~\cite{sun2019bert4rec} uses BERT-like mask language modeling to learn user/item embeddings, integrated with a bidirectional self-attention mechanism,  for recommendations.
  \item $S^3Rec$~\cite{zhou2020s3} extends BERT4Rec by adding auxiliary tasks, such as item attribute prediction to enhance MLM, which can integrate content features for self-supervised learning.
  \item UniSRec~\cite{hou2022towards} leverages item description texts to learn transferable sequence representations across different domains, employing a lightweight architecture with contrastive pre-training tasks for robust performance.
  \item FDSA~\cite{zhang2019feature} enhances prediction accuracy by not only considering item-level transition patterns but also integrating and weighing heterogeneous item features to capture both explicit and implicit feature-level sequences.
  \item SASRec~\cite{kang2018self} captures long-term user behaviors by selectively focusing on relevant past actions.
  \item GRU4Rec~\cite{hidasi2015session} focuses on short session data where traditional matrix factorization fails and demonstrates significant improvements over conventional item-to-item methods.
  \item LightGCN~\cite{he2020lightgcn} ignores feature transformation and nonlinear activation to enhance training efficiency and recommendation performance.
  \item DWSRec~\cite{zhang2024dual} is a sequential recommendation method that relies solely on pre-trained text embeddings and introduces a dual-view whitening strategy to enhance their effectiveness.
  \item HSTU~\cite{zhai2024actions} is a generative recommendation architecture that reformulates recommendation as a sequential transduction task, achieving superior accuracy and scalability on large-scale, high-cardinality data and demonstrating power-law scaling similar to foundation models like GPT-3.
  \item LLMRec~\cite{wei2024llmrec} is a graph augmentation framework that uses large language models to enrich user-item interaction graphs through edge reinforcement, item attribute enhancement, and user profiling.
  \item RecMind~\cite{wang2023recmind} is an LLM-powered autonomous recommender agent that performs zero-shot personalized recommendation by planning with external tools and knowledge.
\end{itemize}

% \vspace{-0.1cm}
\subsubsection{Hyperparameters and Settings}
We conducted experiments using  GPT-2 as the base model. We set the maximum input length to 1024, the token embedding dimension to 768, and the vocabulary length of natural language tokens to 50257. 
For \textbf{Equation~\eqref{eq:r_path}}, $\delta$ is set to 0.9.
In the pre-training stage, we first trained 10 epochs using crowd contextual data to optimize LLM and then trained 100 rounds using user-item interaction data. 
In the fine-tuning stage, we used 50 epochs for the recommendation-oriented fine-tuning of LLM.

% \vspace{-0.1cm}
\subsubsection{Experimental Environment}
All experiments were conducted on Ubuntu 22.04.3 LTS OS, Intel(R) Core(TM) i9-13900KF CPU, with the framework of Python 3.11.5 and PyTorch 2.0.1. All data are computed on an NVIDIA GeForce RTX 4090 GPU, which features 24,576 MiB of memory with CUDA version 12.2.

% \vspace{-0.2cm}
\subsection{Experimental Results}
% \vspace{-0.1cm}

\subsubsection{Overall Comparison}
This experiment aims to answer: {\itshape Can our model really generate more accurate recommendation results through the natural language processing method?}
We compared our model with several baseline models on various Amazon datasets. 
The baseline models used for comparison include ID-based and Attention-based methods. Our model was tested on the same dataset as these baseline models to ensure fairness and accuracy in the comparison.
The experimental results are shown in \textbf{Table~\ref{tab:AmazonComparison}}, and our model performs well in seven Amazon datasets. Recall@20, Recall@40, and NDCG@100 are superior to the baseline models. This indicates that LLMs have strong capabilities in understanding text and capturing user preferences and needs, thereby promoting the accuracy of recommendations.
Overall, the experimental results support our hypothesis that our model can generate more accurate recommendations through the graph-attentive LLM. This discovery is important for research and the practical application of recommender systems.

We also observe that some baselines perform better than others on certain datasets. For example, methods like UniSRec and FDSA perform well on datasets such as AM-Luxury and AM-Scientific. These models benefit from rich sequential user interactions and are optimized for capturing fine-grained temporal or semantic patterns. In contrast, traditional collaborative filtering methods (e.g., Multi-VAE, LightGCN) tend to underperform on such datasets due to their limited capacity for encoding semantic context. Our model shows consistent improvements across datasets by leveraging both textual semantics and graph connectivity in a unified prompt-driven manner.

In addition, we found that some models, including ours, are more sensitive to the characteristics of different datasets. For instance, AM-Food and AM-Instruments are relatively sparse and contain shorter user sequences. Models like SASRec and GRU4Rec, which depend on longer sequences, are affected more on these datasets. Our model maintains strong performance in these cases, as it integrates higher-order user-item relationships via graph structure and uses prompt-based representations to compensate for interaction sparsity. These observations highlight the importance of incorporating both structure and semantics in low-data or sparse interaction scenarios.

\begin{table*}[htb]
  \centering
  \caption{Comparison Between Our Model and Baselines on Three Amazon Review Datasets.}
  \label{tab:AmazonComparison}
  \resizebox{\textwidth}{!}{
  \begin{tabular}{llcccccccccccccc}
    \toprule
    Dataset & Metric & Multi-VAE & MD-CVAE & LightGCN & BERT4Rec & $S^3$Rec & UniSRec & FDSA & SASRec & GRU4Rec & DWSRec & HSTU & LLMRec & RecMind & Ours \\
    \midrule
    \multirow{3}{*}{\textbf{AM-Beauty}} & Recall@20 & 0.1295 & 0.1472 & 0.1429 & 0.1126 & 0.1354 & 0.1462 & 0.1447 & 0.1503 & 0.0997 & 0.1510 & 0.1546 & 0.1508 & 0.1347 & \textbf{0.1590} \\
    & Recall@40 & 0.1720 & 0.2058 & 0.1967 & 0.1677 & 0.1789 & 0.1898 & 0.1875 & 0.2018 & 0.1528 & 0.1985 & 0.2104 & 0.2018 & 0.1874 & \textbf{0.2177} \\
    & NDCG@100 & 0.0835 & 0.0871 & 0.0890 & 0.0781 & 0.0867 & 0.0907 & 0.0834 & 0.0929 & 0.0749 & 0.0971 & 0.0973 & 0.0927 & 0.0846 & \textbf{0.1029} \\
    \midrule
    \multirow{3}{*}{\textbf{AM-Toys}} & Recall@20 & 0.1076 & 0.1107 & 0.1096 & 0.0853 & 0.1064 & 0.1110 & 0.0972 & 0.0869 & 0.0657 & 0.1307 & 0.912 & 0.1207 & 0.1126 & \textbf{0.1349} \\
    & Recall@40 & 0.1558 & 0.1678 & 0.1558 & 0.1375 & 0.1524 & 0.1457 & 0.1268 & 0.1146 & 0.0917 & 0.1749 & 0.1208 & 0.1639 & 0.1564& \textbf{0.1873} \\
    & NDCG@100 & 0.0781 & 0.0812 & 0.0775 & 0.0532 & 0.0665 & 0.0638 & 0.0662 & 0.0525 & 0.0439 & 0.0784 & 0.0569 & 0.0672 & 0.0584 & \textbf{0.0876} \\
    \midrule
    \multirow{3}{*}{\textbf{AM-Sports}} & Recall@20 & 0.0659 & 0.0714 & 0.0677 & 0.0521 & 0.0616 & 0.0714 & 0.0681 & 0.0541 & 0.0720 & 0.0753 & 0.0631 & 0.0701 & 0.0683 & \textbf{0.0764} \\
    & Recall@40 & 0.0975 & 0.1180 & 0.0973 & 0.0701 & 0.0813 & 0.1143 & 0.0866 & 0.0739 & 0.1086 & 0.0960 & 0.0868 & 0.1183 & 0.1147 & \textbf{0.1240} \\
    & NDCG@100 & 0.0446 & 0.0514 & 0.0475 & 0.0305 & 0.0438 & 0.0504 & 0.0475 & 0.0361 & 0.0498 & 0.0484 & 0.0399 & 0.0498 & 0.0511 & \textbf{0.0535} \\
    \midrule
    \multirow{3}{*}{\textbf{AM-Luxury}} & Recall@20 & 0.2306 & 0.2771 & 0.2514 & 0.2076 & 0.2241 & \textbf{0.3091} & 0.2759 & 0.2550 & 0.2126 & 0.2524 & 0.2779 & 0.2761 & 0.2879 & 0.3066 \\
    & Recall@40 & 0.2724 & 0.3206 & 0.3004 & 0.2404 & 0.2672 & \textbf{0.3675} & 0.3176 & 0.3008 & 0.2522 & 0.2876 & 0.3206 & 0.3219 & 0.3351 & 0.3441 \\
    & NDCG@100 & 0.1697 & 0.2064 & 0.1947 & 0.1617 & 0.1542 & 0.2010 & 0.2107 & 0.1965 & 0.1623 & 0.1476 & 0.2177 & 0.2017 & 0.2049 & \textbf{0.2331} \\
    \midrule
    \multirow{3}{*}{\textbf{AM-Scientific}} & Recall@20 & 0.1069 & 0.1389 & 0.1385 & 0.0871 & 0.1089 & \textbf{0.1492} & 0.1188 & 0.1298 & 0.0849 & 0.1096 & 0.1401 & 0.1409 & 0.1274 & 0.1480 \\
    & Recall@40 & 0.1483 & 0.1842 & 0.1857 & 0.1160 & 0.1541 & \textbf{0.1954} & 0.1547 & 0.1776 & 0.1204 & 0.1360 & 0.1748 & 0.1839 & 0.1651 & 0.1908 \\
    & NDCG@100 & 0.0766 & 0.0872 & 0.0834 & 0.0606 & 0.0715 & 0.1056 & 0.0846 & 0.0864 & 0.0594 & 0.0645 & 0.0927 & 0.0978 & 0.0873 & \textbf{0.1072} \\
    \midrule
    \multirow{3}{*}{\textbf{AM-Instruments}} & Recall@20 & 0.1096 & 0.1398 & 0.1195 & 0.1183 & 0.1352 & 0.1684 & 0.1382 & 0.1483 & 0.1271 & 0.1057 & 0.1563 & 0.1322 & 0.1539 & \textbf{0.1698} \\
    & Recall@40 & 0.1628 & 0.1743 & 0.1575 & 0.1531 & 0.1767 & 0.2239 & 0.1787 & 0.1935 & 0.1660 & 0.1423 & 0.2103 & 0.1727 & 0.2093 & \textbf{0.2265} \\
    & NDCG@100 & 0.0735 & 0.1040 & 0.0985 & 0.0922 & 0.0894 & 0.1075 & 0.1080 & 0.0934 & 0.0998 & 0.0682 & 0.1074 & 0.0937 & 0.1008 & \textbf{0.1312} \\
    \midrule
    \multirow{3}{*}{\textbf{AM-Food}} & Recall@20 & 0.1062 & 0.1170 & 0.1149 & 0.1036 & 0.1157 & 0.1423 & 0.1099 & 0.1171 & 0.1140 & 0.1341 & 0.1204 & 0.1347 & 0.1194 & \textbf{0.1438} \\
    & Recall@40 & 0.1317 & 0.1431 & 0.1385 & 0.1284 & 0.1456 & 0.1661 & 0.1317 & 0.1404 & 0.1389 & 0.1618 & 0.1477 & 0.1564 & 0.1399 & \textbf{0.1673} \\
    & NDCG@100 & 0.0727 & 0.0863 & 0.0853 & 0.0835 & 0.0926 & 0.1024 & 0.0904 & 0.0942 & 0.0910 & 0.0823 & 0.0929 & 0.0993 & 0.0783 & \textbf{0.1119} \\
    \bottomrule
  \end{tabular}
  }
\end{table*}

\subsubsection{Ablation Studies}
This experiment aims to answer: {\itshape How essential are each component's contributions to our model?}
To answer this question, we designed the following experimental baselines:

\begin{itemize}
  \item LLM-NoPretrain removes the use of pre-trained models and starts training models from scratch to evaluate the impact of pre-training steps on performance.
  \item LLM-NoFineTune directly uses the model and embedding for recommendation tasks after pre-training without any fine-tuning steps.
  \item LLM-NoGKIA does not integrate graph knowledge into attention mechanisms to evaluate the contribution of incorporating graph structure information into the model's performance.
  \item LLM-NoGHIP does not include graph or historical information to embed prompts for pre-training but only uses simple users' review information to evaluate the impact of complex prompts on model performance.
\end{itemize}

We pre-trained and fine-tuned each baseline model separately, and then compared it with our complete model. These pre-training and fine-tuning experimental settings are consistent and conducted on the same dataset to ensure the comparability of results.
\textbf{Table~\ref{tab:AblationStudy}} shows each component's specific contribution to the model's overall performance. For example, the performance of LLM-NoPretrain is significantly lower than that of the complete model. This implies that using recommendation-related graph data and natural language data for pre-training plays a crucial role in improving model performance. 
Similarly, the results of LLM-NoFineTune demonstrate the importance of fine-tuning. 
Subsequently, by comparing the performance of LLM-NoGKIA, and LLM-NoGHIP with that of the complete model, we find that the addition of graph connection information in attention calculation and complex prompts containing second-order relationships is crucial for improving the performance of recommender systems.

\begin{table*}[htb]
  \centering
  \caption{Ablation Study Results.}
  \label{tab:AblationStudy}
  \resizebox{0.7\textwidth}{!}{
  \begin{tabular}{llccccc}
    \toprule
    Dataset & Metric & LLM-NoPretrain & LLM-NoFineTune & LLM-NoGKIA & LLM-NoGHIP & Ours \\
    \midrule
    \multirow{3}{*}{\textbf{AM-Beauty}} & Recall@20 & 0.0464 & 0.0441 & 0.1225 & 0.1267 & \textbf{0.1590} \\
    & Recall@40 & 0.0709 & 0.0691 & 0.1665 & 0.1799 & \textbf{0.2177} \\
    & NDCG@100 & 0.0339 & 0.0323 & 0.0790 & 0.0827 & \textbf{0.1029} \\
    \midrule
    \multirow{3}{*}{\textbf{AM-Toys}} & Recall@20 & 0.0477 & 0.0580 & 0.0896 & 0.0858 & \textbf{0.1349} \\
    & Recall@40 & 0.0689 & 0.1003 & 0.1272 & 0.1179 & \textbf{0.1873} \\
    & NDCG@100 & 0.0330 & 0.0481 & 0.0612 & 0.0594 & \textbf{0.0876} \\
    \midrule
    \multirow{3}{*}{\textbf{AM-Sports}} & Recall@20 & 0.0449 & 0.0394 & 0.0555 & 0.0558 & \textbf{0.0764} \\
    & Recall@40 & 0.0719 & 0.0613 & 0.0846 & 0.0830 & \textbf{0.1240} \\
    & NDCG@100 & 0.0322 & 0.0278 & 0.0391 & 0.0379 & \textbf{0.0535} \\
    \midrule
    \multirow{3}{*}{\textbf{AM-Luxury}} & Recall@20 & 0.1872 & 0.1885 & 0.2474 & 0.2679 & \textbf{0.3066} \\
    & Recall@40 & 0.2233 & 0.2254 & 0.2880 & 0.3028 & \textbf{0.3441} \\
    & NDCG@100 & 0.1223 & 0.1235 & 0.1834 & 0.2065 & \textbf{0.2331} \\
    \midrule
    \multirow{3}{*}{\textbf{AM-Scientific}} & Recall@20 & 0.0708 & 0.0668 & 0.1383 & 0.1206 & \textbf{0.1480} \\
    & Recall@40 & 0.1037 & 0.0960 & 0.1822 & 0.1575 & \textbf{0.1908} \\
    & NDCG@100 & 0.0568 & 0.0465 & 0.0940 & 0.0810 & \textbf{0.1072} \\
    \midrule
    \multirow{3}{*}{\textbf{AM-Instruments}} & Recall@20 & 0.0766 & 0.0727 & 0.1387 & 0.1426 & \textbf{0.1698} \\
    & Recall@40 & 0.1004 & 0.0948 & 0.1741 & 0.1779 & \textbf{0.2265} \\
    & NDCG@100 & 0.0500 & 0.0478 & 0.1042 & 0.1044 & \textbf{0.1312} \\
    \midrule
    \multirow{3}{*}{\textbf{AM-Food}} & Recall@20 & 0.0224 & 0.0204 & 0.1275 & 0.1264 & \textbf{0.1438} \\
    & Recall@40 & 0.0299 & 0.0274 & 0.1559 & 0.1487 & \textbf{0.1673} \\
    & NDCG@100 & 0.0153 & 0.0141 & 0.0898 & 0.0963 & \textbf{0.1119} \\
    \bottomrule
  \end{tabular}
  }
\end{table*}

\subsubsection{Study on Different Pre-training Prompt Structures}
This experiment aims to answer: {\itshape What is the contribution of pre-training text data that integrates second-order relationships and item background information to recommendation models?}
To answer this question, we chose the AM-Toys dataset and designed the following experimental models:
\begin{itemize}
  \item Entire Prompt Model: A complete model that includes text data with second-order relationships and items' background information.
  \item Without 2-Order: The model does not contain 2-order relationship information.
  \item Without Item: The model does not contain items' background information.
  \item Without Prompt: The model does not contain any second-order relationship information or item background information.
\end{itemize}

By comparing the performances of these models, we quantified the impact of the second-order relationships and the background information of items on recommendation accuracy.

\begin{figure}[htbp]
\centering
\includegraphics[width=0.6\textwidth]{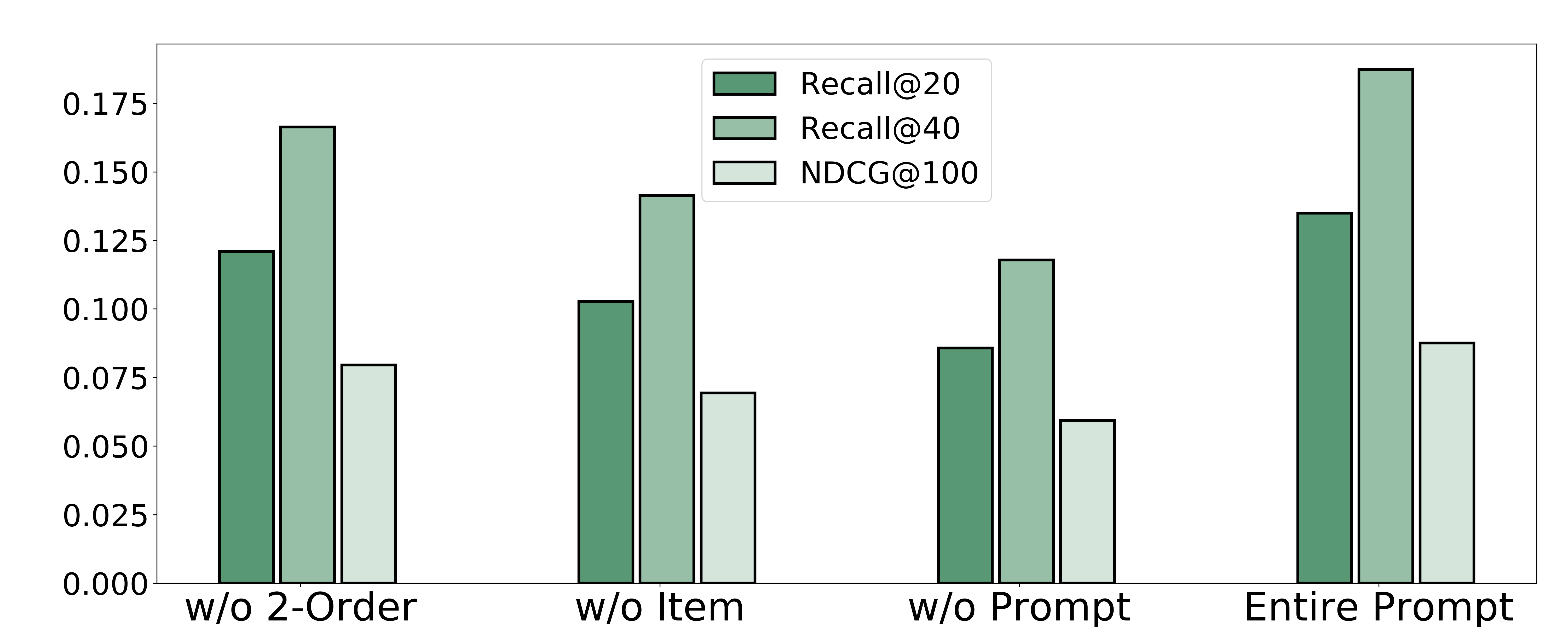}
\caption{Results of Different Prompt Structures.}
% \vspace{-0.6cm}
\label{fig:toysPrompt}
\end{figure}

Figure ~\ref{fig:toysPrompt} shows that the model that uses prompt sentences of complete information (with the second-order relationship) performs best over all the performance indicators. 
The performances of the "Without second-order relationship" model are lower than that of the complete model. As can be seen, second-order relationship information is an essential component of graph connectivity. Similarly, the "Without Item" model performs poorly, highlighting the importance of natural language background information in enhancing recommender systems.

\subsubsection{Study on Different Historical Interaction Length}
This experiment aims to answer: {\itshape How does the historical interaction length influence the recommendation performance?}
Here, we employ different numbers of historical interactions in the fine-tuning prompts. Because in real-world scenarios, there would be numerous historical interactions. Selecting appropriate numbers will find a balance between cache memory and performance. 

\begin{figure}[htbp]
\centering
\includegraphics[width=0.45\textwidth]{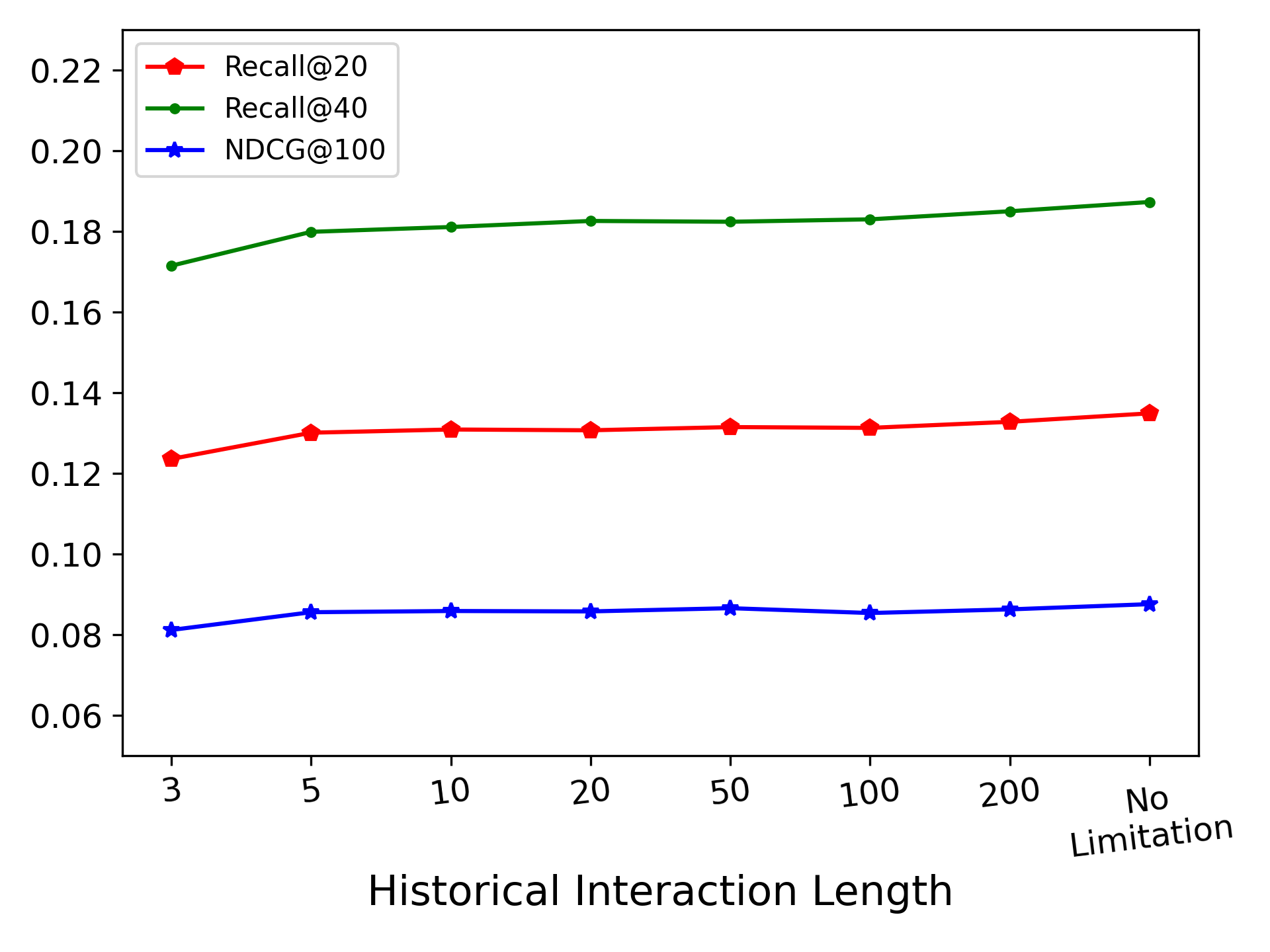}
\caption{Results of Different Historical Length.}
% \vspace{-0.6cm}
\label{fig:toyshislen}
\end{figure}

The results show that the length of historical events does affect the performance. However, the difference between different parameters is not significant. Therefore, when employing the system, we have more flexibility to choose the appropriate historical length.

\subsubsection{Study on Different Attention Injection Ways}
This experiment aims to answer: {\itshape Is the connection information in the attention calculation process of the GPT-2 model really that important?}
To answer this question, we used the AM-Toys and AM-Beauty datasets. The experimental design included three different attention mechanisms: 
\begin{itemize}
  \item Reasonable Injection: Injecting meaningful connection information into the attention mechanism.
  \item Meaningless Injection: Set all connection information of the attention mechanism to 1, without considering actual connection strength or relationships.
  \item Normal Attention: Maintain the normal attention mechanism of the GPT-2 model without any injection.
\end{itemize}

\begin{figure}[htp]
        \centering
        \subfigure[AM-Beauty]{\includegraphics[width=0.3\textwidth]{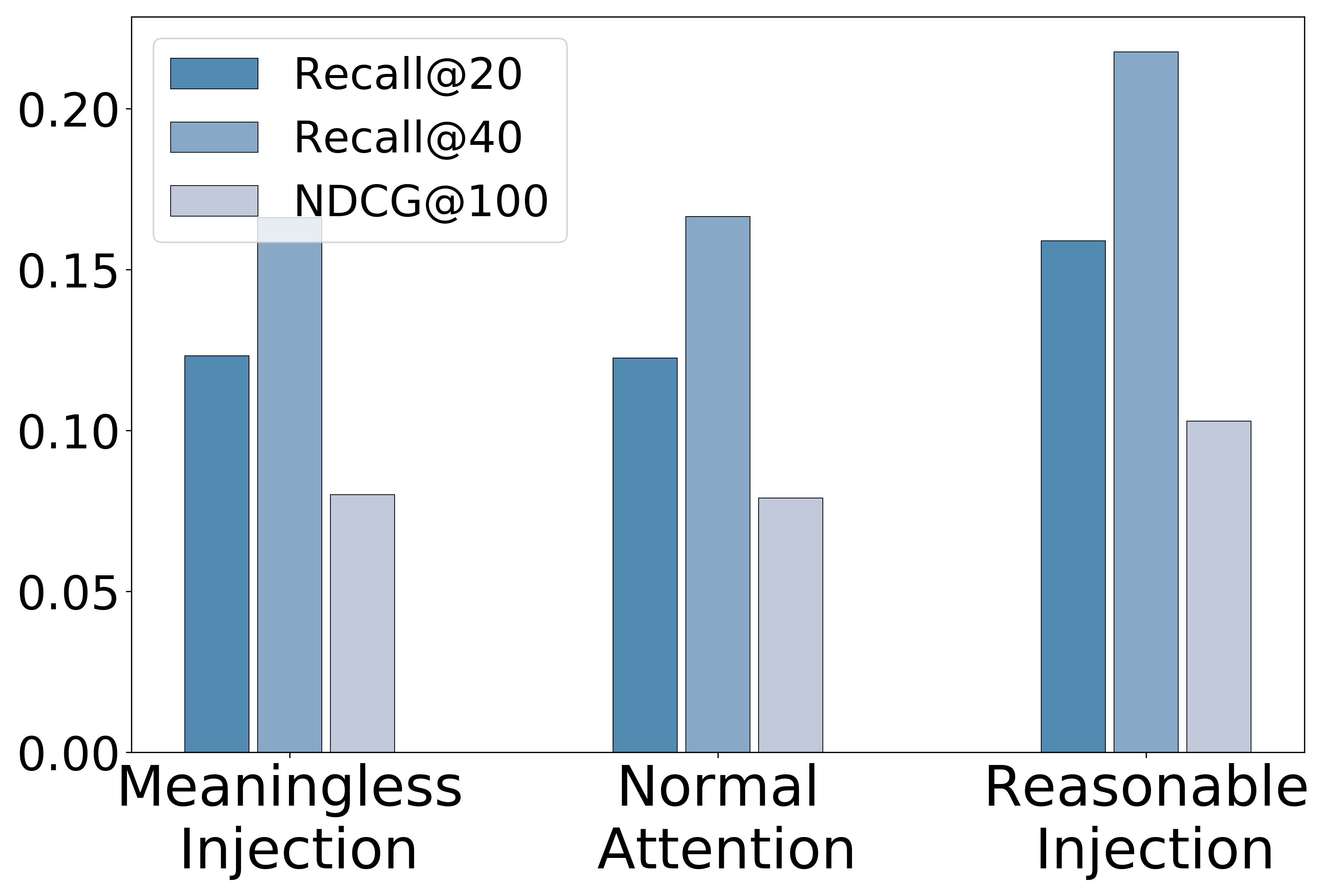}}
        \subfigure[AM-Toys]{\includegraphics[width=0.3\textwidth]{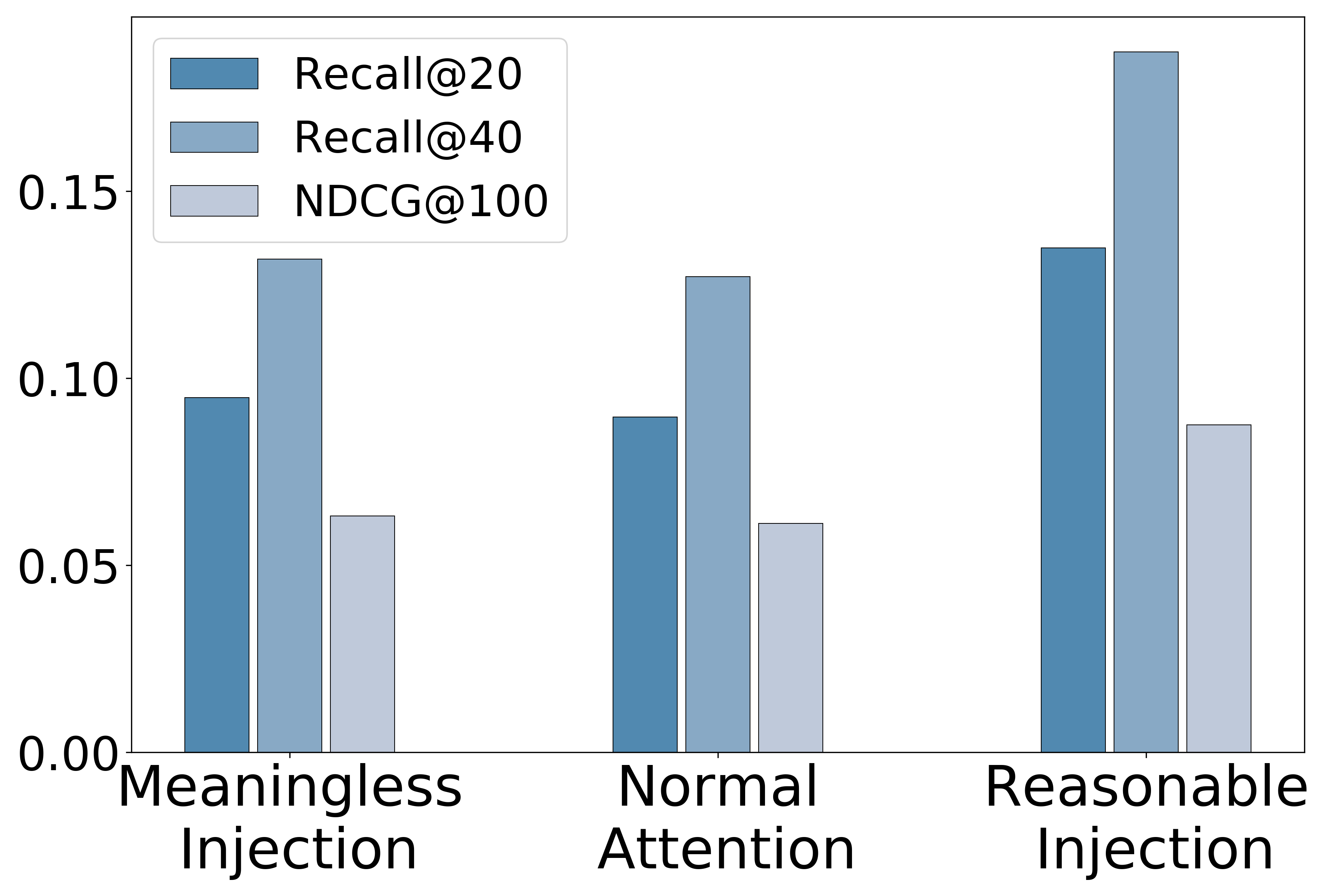}}
        \subfigure[AM-Sports]{\includegraphics[width=0.3\textwidth]{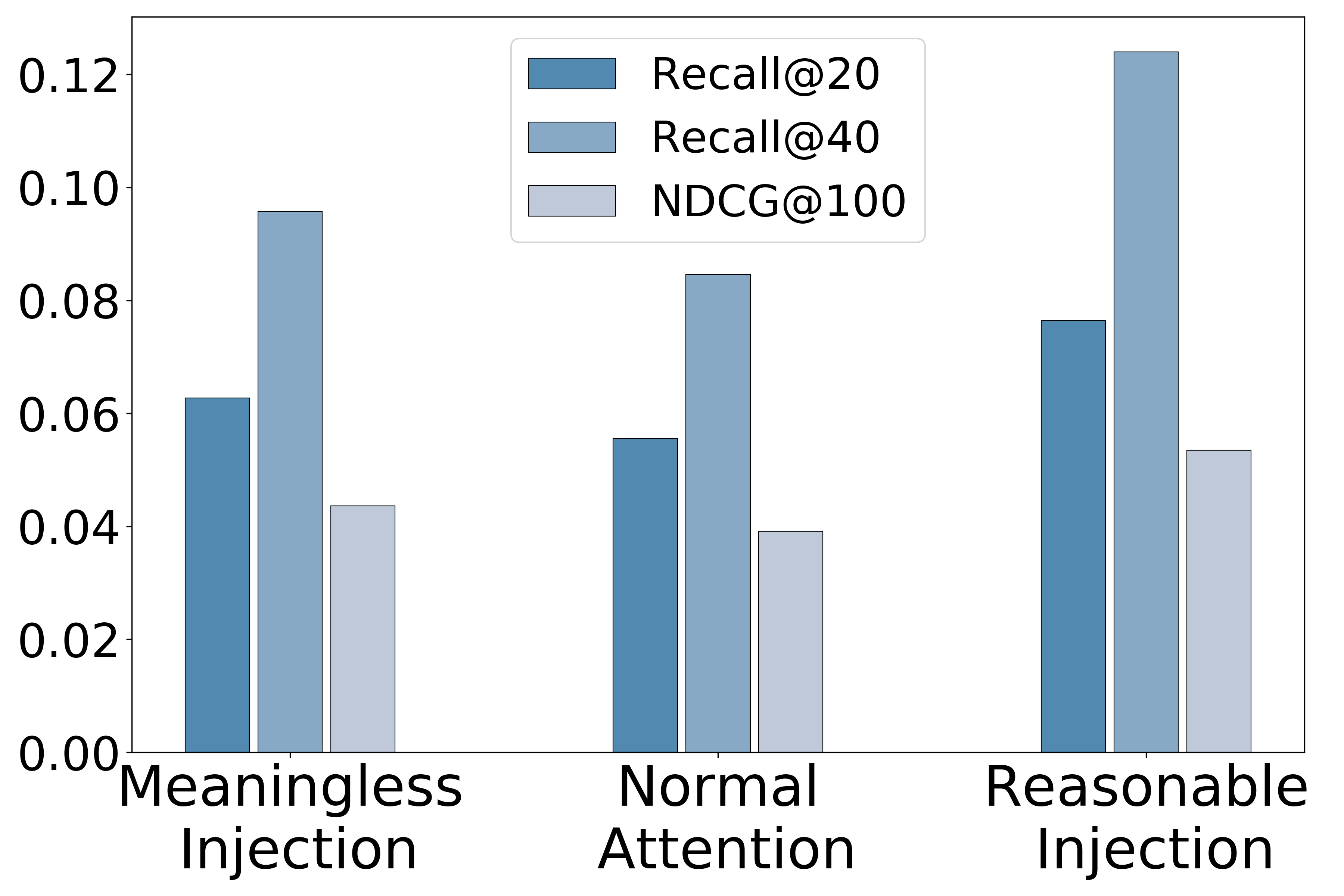}}
        % \vspace{-0.3cm}
        \caption{Results of Different Attention Injection Ways.}
        % \vspace{-0.5cm}
        \label{fig:beautytoysAttention}
\end{figure}

% \begin{figure}[htbp]
% \centering
% \includegraphics[width=0.5\textwidth]{figures/beauty_toys_attention.png}
% \caption{Results of Different Attention Injection Ways.}
% \label{fig:beautytoysAttention}
% \end{figure}

\textbf{Fig.~\ref{fig:beautytoysAttention}} shows that the model using our graph attentive LLM method exhibits the best performance.  Our method not only considers the direct connections between nodes but also the spatial relationships (i.e., the shortest connected path) between nodes in the graph. 

\begin{figure}[htbp]
\centering
\includegraphics[width=0.45\textwidth]{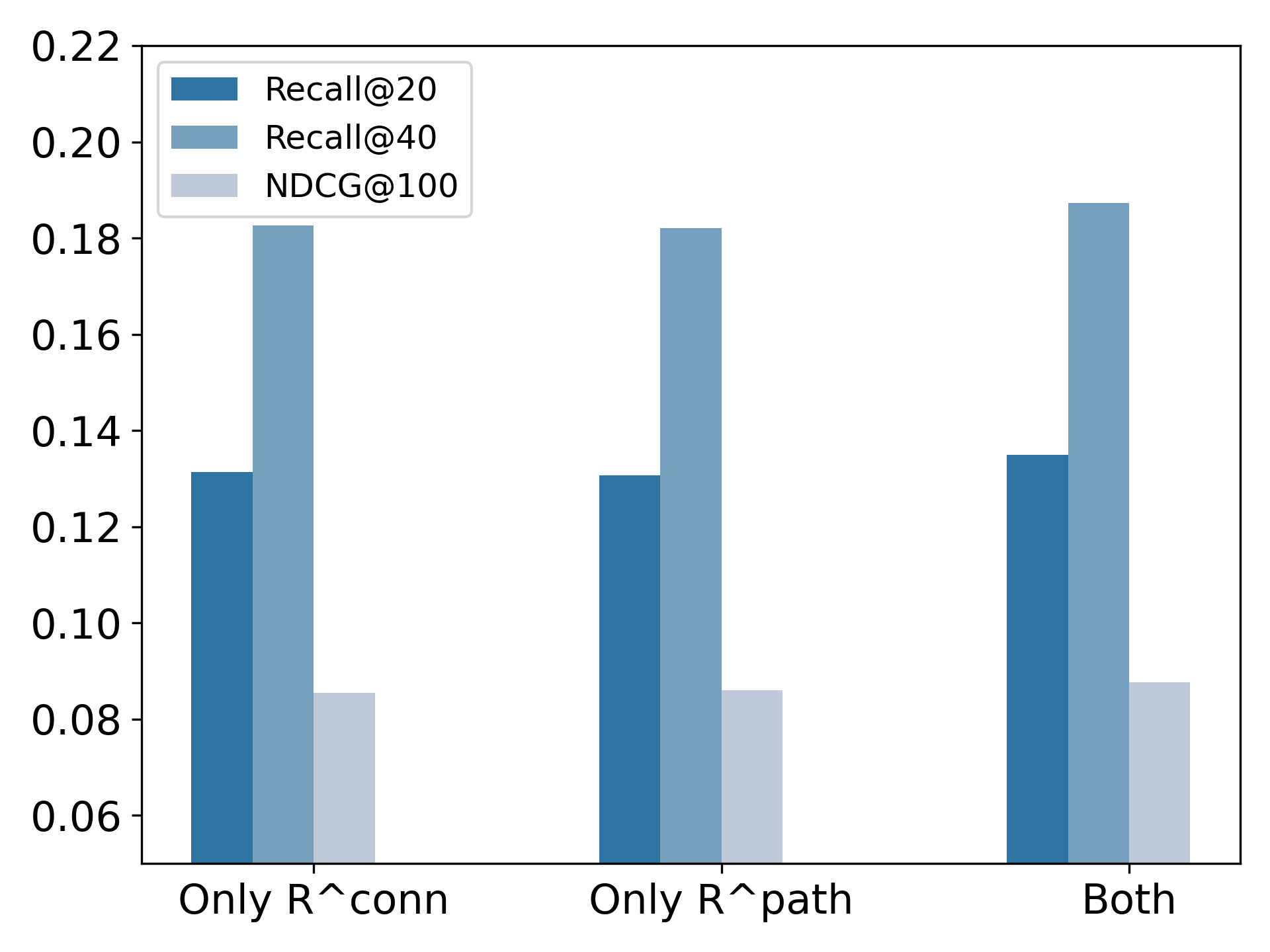}
\caption{Results of Two Parts of $R$.}
% \vspace{-0.3cm}
\label{fig:toysR2parts}
\end{figure}

We compared our method with regular attention mechanisms, and the experimental results clearly support this point. 
To avoid bias that may arise from adding input only between user/project tokens, we introduced a comparison with fixed additive attention. We found that simply adding fixed connection information to attention calculation for nodes in the graph is not effective. It is truly effective to include information that reflects the actual relationships between nodes.
Meanwhile, our experiments on the AM-Toys dataset, with results shown in \textbf{Fig.~\ref{fig:toysR2parts}} show that both direct $R^{\text{conn}}$ and indirect connective information $R^{\text{path}}$ contribute to the performance.

\subsubsection{Study of Parameters}
\begin{figure}[htp]
        \centering
        \subfigure[Recall@20]{\includegraphics[width=0.3\textwidth]{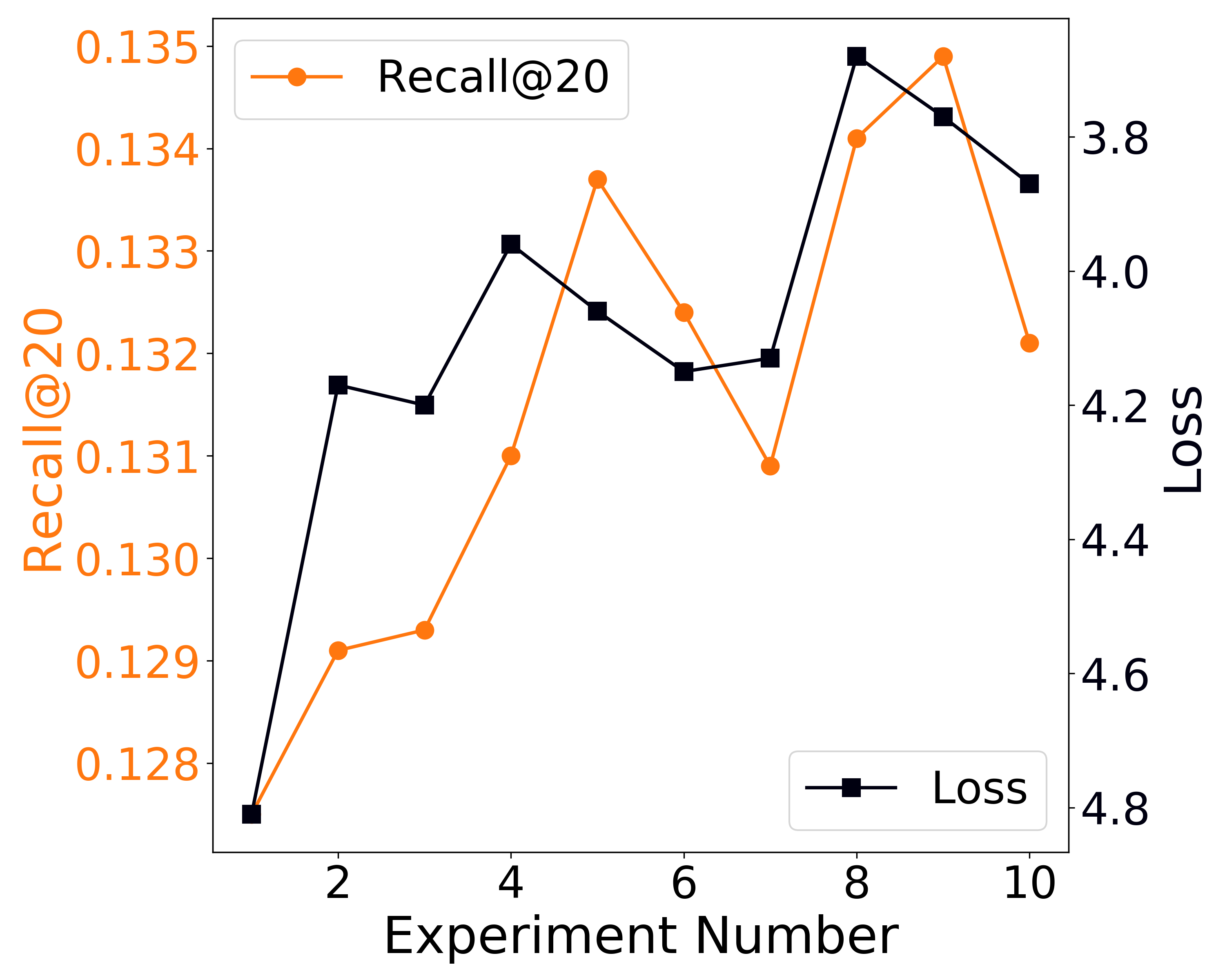}}
        \subfigure[Recall@40]{\includegraphics[width=0.3\textwidth]{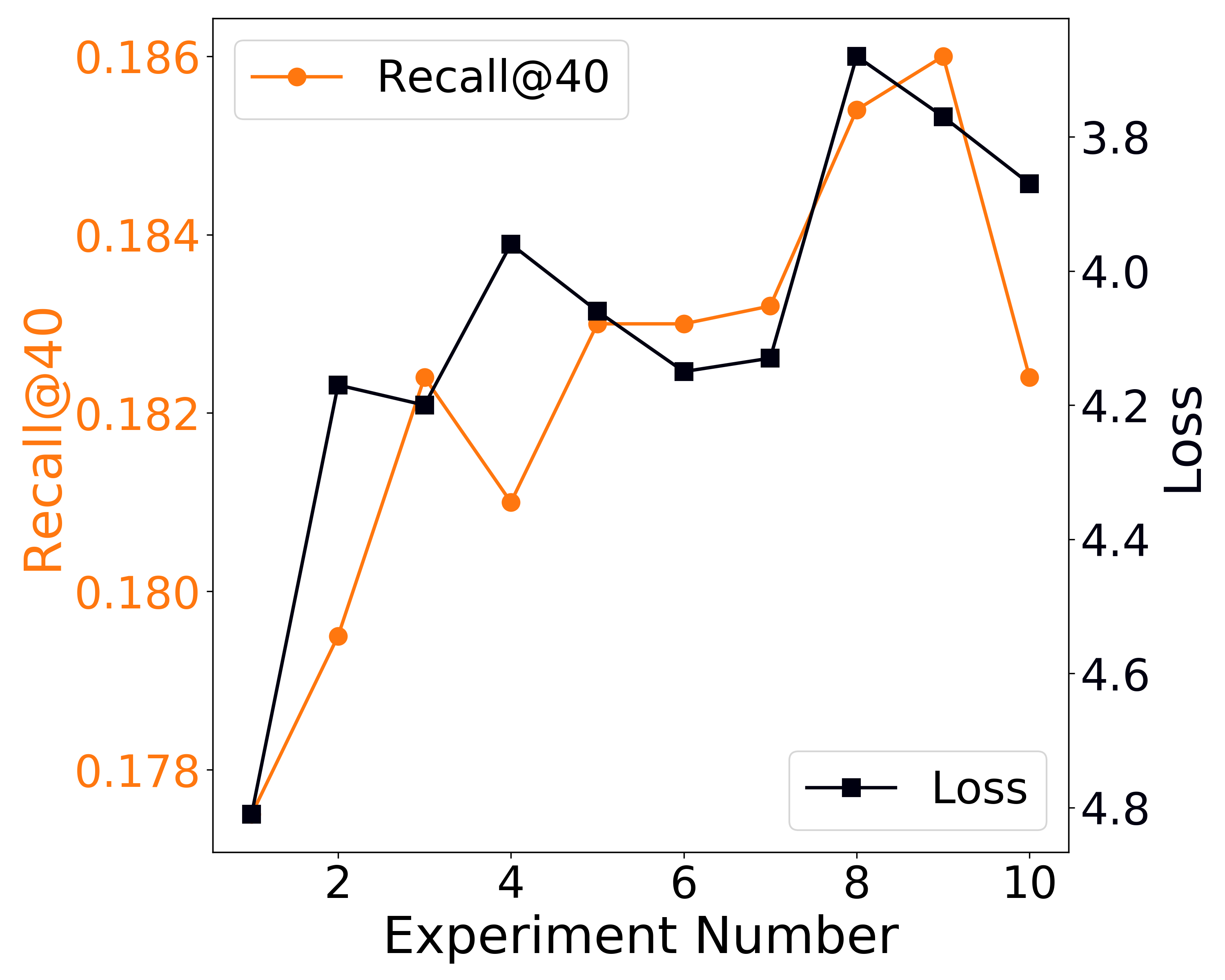}}
        \subfigure[NDCG@100]{\includegraphics[width=0.3\textwidth]{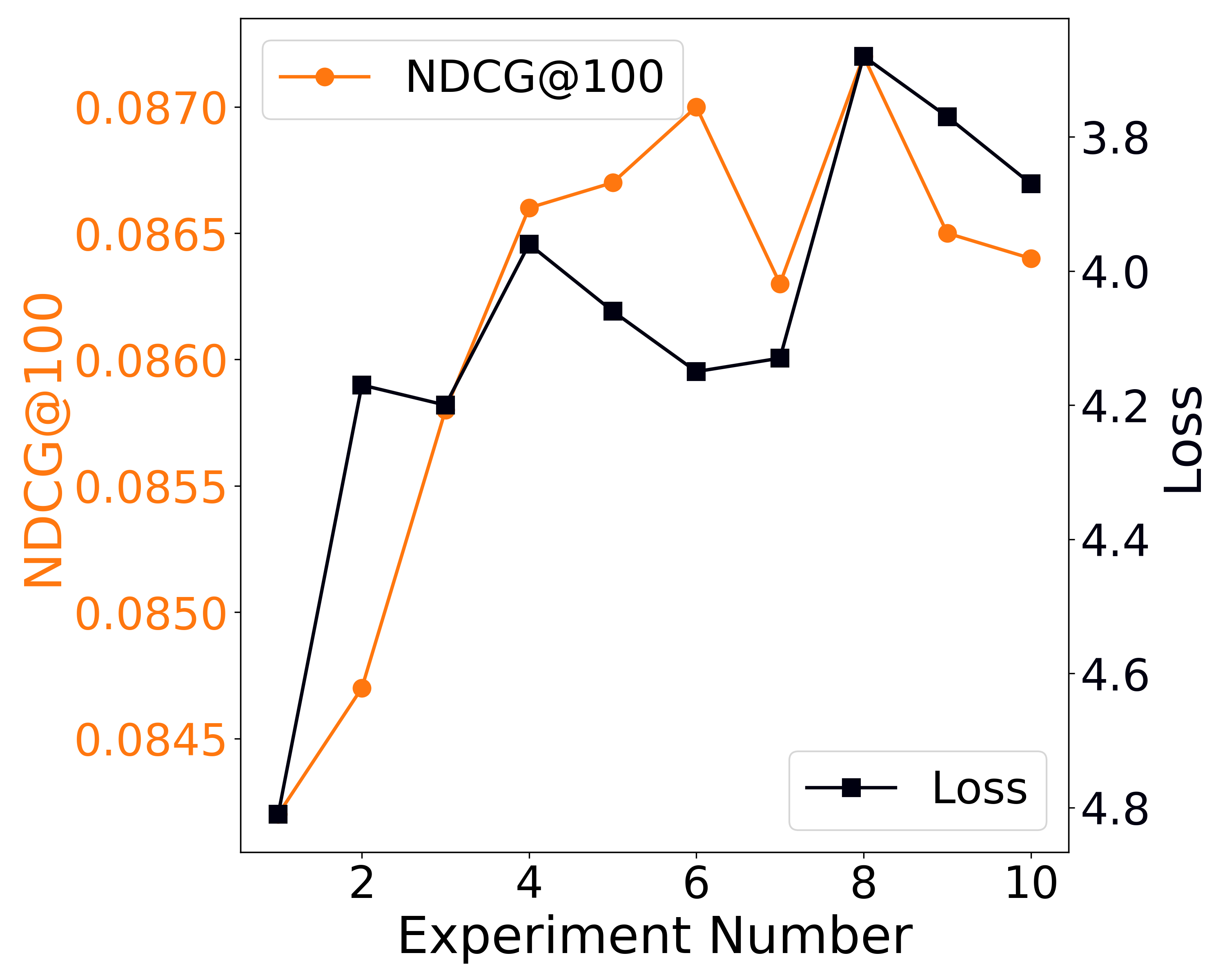}}
        % \vspace{-0.3cm}
        \caption{Results of Different Training Epochs.}
        % \vspace{-0.5cm}
        \label{fig:toysepoch}
\end{figure}

This experiment aims to answer: {\itshape Can we ensure consistency between our pre-training and fine-tuning tasks?}
We conducted experiments on the AM-Toys dataset to analyze the performance alignment between the pre-training task and the fine-tuning task. 
We used the results of the first 10 pre-training epochs and the corresponding loss function. 
Then, we fine-tuned the pre-trained model to obtain evaluation metrics.  
We compared the 3 metrics, Recall@20, Recall@40, and NDCG@100 with the loss function.
\textbf{Fig.~\ref{fig:toysepoch}} shows the trend of changes in the 3 metrics is consistent with the trend of changes in loss functions. This indicates that our pre-training task and fine-tuning task are well-aligned, and our prompt construction method can provide rich information for subsequent recommendation tasks.

\subsubsection{Study of Different LLM backbones}

To explore the generalizability and robustness of our method across different large language model (LLM) architectures, we further conduct experiments using \textbf{BART} as the backbone, in addition to GPT-2. Unlike GPT-2, which is a decoder-only transformer, BART is a sequence-to-sequence model that integrates both transformer encoder and decoder components. This architectural difference allows BART to potentially better capture bidirectional contextual information, which may influence recommendation outcomes.

\begin{figure}[htbp]
  \centering
  \subfigure[Standard Attention]{
    \includegraphics[width=0.35\textwidth]{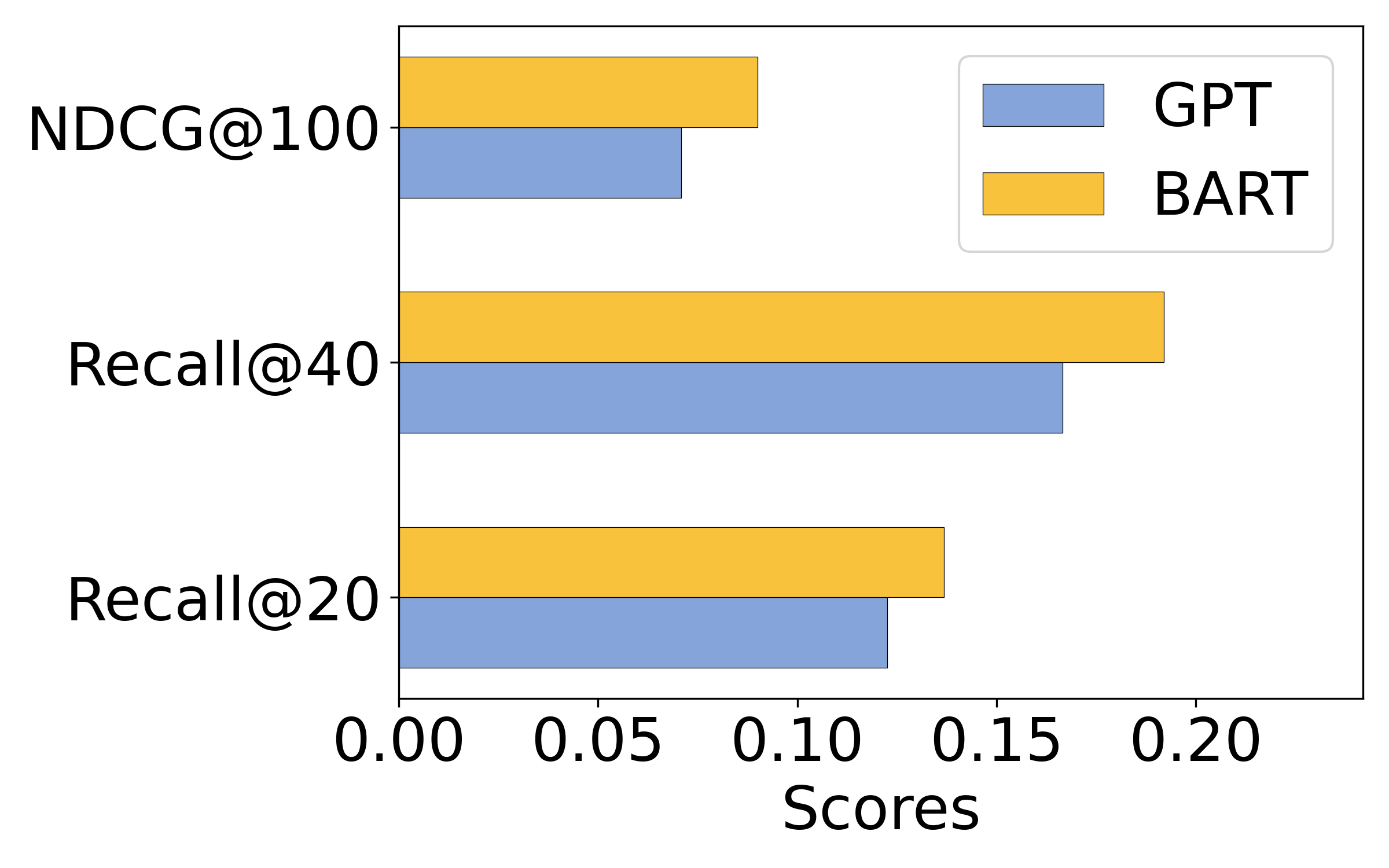}
    \label{fig:gpt}
  }
  % \hfill
  \subfigure[Graph Information Integrated]{
    \includegraphics[width=0.35\textwidth]{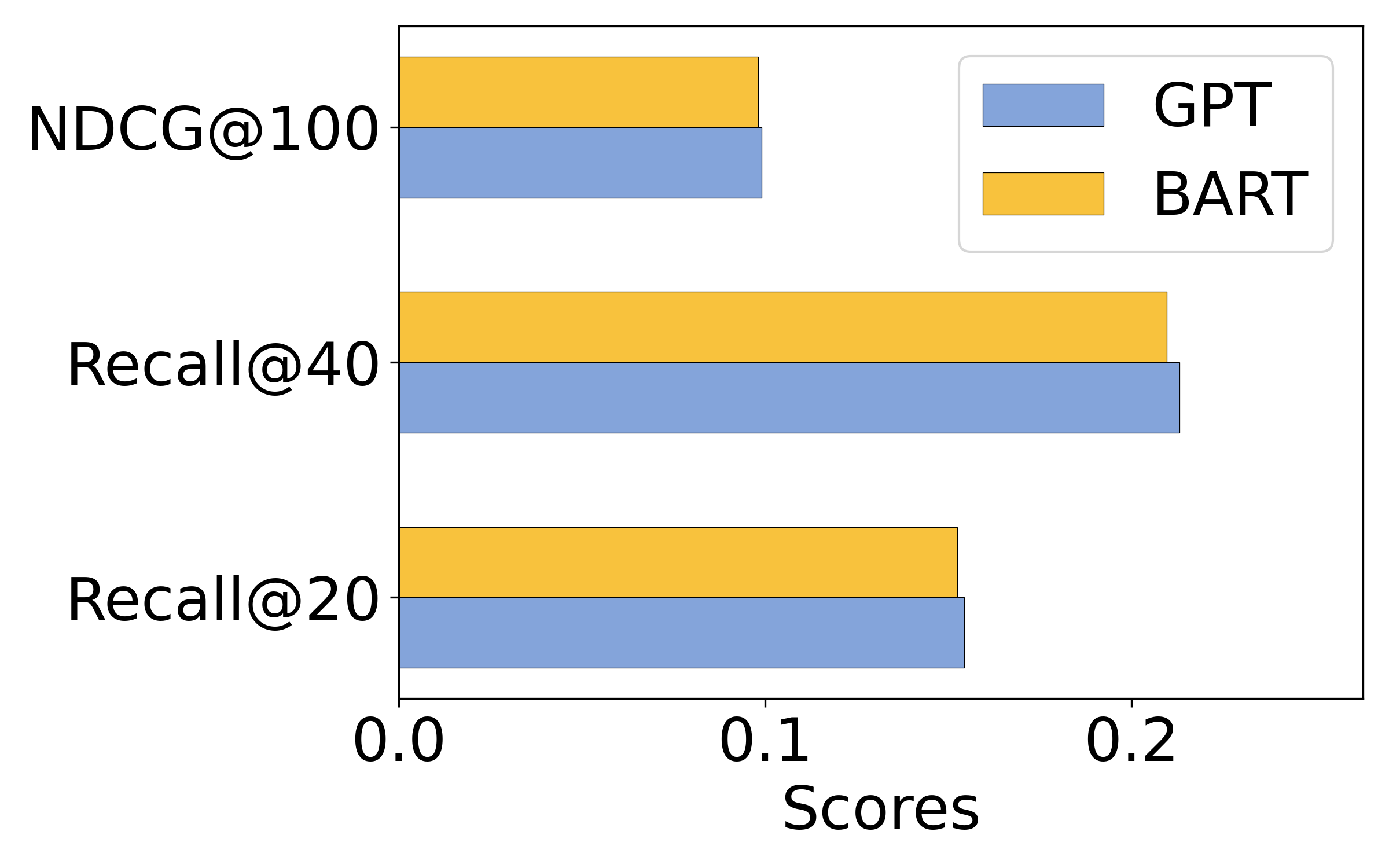}
    \label{fig:bart}
  }
  \caption{Comparison between Different LLM Backbones.}
  \label{fig:llm_backbones_comparison}
\end{figure}

As shown in \textbf{Fig.~\ref{fig:llm_backbones_comparison}}, we compare the performance of GPT and BART under two settings: (a) with \textit{standard attention}, and (b) with \textit{graph information integration}. The models are evaluated using NDCG@100, Recall@40, and Recall@20.

From the results, we observe that:
\begin{itemize}
    \item Under \textit{standard attention}, the BART-based model performs marginally better than GPT in all three metrics.
    \item However, with \textit{graph information integrated}, the performance gap diminishes, and GPT slightly outperforms BART in Recall@20.
\end{itemize}

These findings indicate that while BART's encoder-decoder structure brings benefits in general scenarios, GPT-based models remain competitive and may perform better when enhanced with graph-structured information. This highlights the adaptability of our framework across different LLMs and the importance of backbone selection based on task characteristics.

\subsubsection{Study of Cold-start Situation}

To evaluate the performance of our model in cold-start scenarios, we conduct additional experiments by simulating users with limited interaction history. Specifically, we randomly reduce each user's training item list by 50\%, which will limit the available data for LLM to learn user preferences. This setup mimics the cold-start condition where users have interacted with only a few items.

\begin{table*}[htb]
  \centering
  \caption{Performance under Normal and Cold-start Settings.}
  \label{tab:ColdStartStudy}
  \resizebox{0.99\textwidth}{!}{
  \begin{tabular}{llcccccc}
    \toprule
    Dataset & Metric & BERT4Rec-Normal & BERT4Rec-Cold & RecMind-Normal & RecMind-Cold & Ours-Normal & Ours-Cold \\
    \midrule
    \multirow{3}{*}{\textbf{AM-Beauty}} & Recall@20 & 0.1126 & 0.0921 & 0.1347 & 0.1206 & 0.1590 & 0.1384 \\
    & Recall@40 & 0.1677 & 0.1439 & 0.1874 & 0.1634 & 0.2177 & 0.1944 \\
    & NDCG@100 & 0.0781 & 0.0631 & 0.0846 & 0.0730 & 0.1029 & 0.0812 \\
    \midrule
    \multirow{3}{*}{\textbf{AM-Toys}} & Recall@20 & 0.0853 & 0.0711 & 0.1126 & 0.0999 & 0.1349 & 0.1274 \\
    & Recall@40 & 0.1375 & 0.1181 & 0.1564 & 0.1392 & 0.1873 & 0.1763 \\
    & NDCG@100 & 0.0532 & 0.0435 & 0.0584 & 0.0506 & 0.0876 & 0.0654 \\
    \midrule
    \multirow{3}{*}{\textbf{AM-Sports}} & Recall@20 & 0.0521 & 0.0451 & 0.0683 & 0.0605 & 0.0764 & 0.0658 \\
    & Recall@40 & 0.0701 & 0.0584 & 0.1147 & 0.1020 & 0.1240 & 0.1102 \\
    & NDCG@100 & 0.0305 & 0.0258 & 0.0511 & 0.0452 & 0.0535 & 0.0502 \\
    \midrule
    \multirow{3}{*}{\textbf{AM-Luxury}} & Recall@20 & 0.2076 & 0.1735 & 0.2879 & 0.2499 & 0.3066 & 0.3044 \\
    & Recall@40 & 0.2404 & 0.1968 & 0.3351 & 0.2860 & 0.3441 & 0.3389 \\
    & NDCG@100 & 0.1617 & 0.1393 & 0.2049 & 0.1832 & 0.2331 & 0.2005 \\
    \midrule
    \multirow{3}{*}{\textbf{AM-Scientific}} & Recall@20 & 0.0871 & 0.0734 & 0.1274 & 0.1112 & 0.1480 & 0.1445 \\
    & Recall@40 & 0.1160 & 0.0989 & 0.1651 & 0.1427 & 0.1908 & 0.1876 \\
    & NDCG@100 & 0.0606 & 0.0517 & 0.0873 & 0.0774 & 0.1072 & 0.1023 \\
    \midrule
    \multirow{3}{*}{\textbf{AM-Instruments}} & Recall@20 & 0.1183 & 0.1009 & 0.1539 & 0.1359 & 0.1698 & 0.1544 \\
    & Recall@40 & 0.1531 & 0.1268 & 0.2093 & 0.1877 & 0.2265 & 0.1943 \\
    & NDCG@100 & 0.0922 & 0.0761 & 0.1008 & 0.0851 & 0.1312 & 0.1163 \\
    \midrule
    \multirow{3}{*}{\textbf{AM-Food}} & Recall@20 & 0.1036 & 0.0863 & 0.1194 & 0.1073 & 0.1438 & 0.1316 \\
    & Recall@40 & 0.1284 & 0.1130 & 0.1399 & 0.1262 & 0.1673 & 0.1546 \\
    & NDCG@100 & 0.0835 & 0.0713 & 0.0783 & 0.0680 & 0.1119 & 0.0996 \\
    \bottomrule
  \end{tabular}
  }
\end{table*}

Table~\ref{tab:ColdStartStudy} shows the results, where we compare the performance of our method with baseline models under both normal and cold-start settings. Our model demonstrates relatively stable performance degradation, suggesting its robustness in data-sparse scenarios. Compared to BERT4Rec and RecMind, our method consistently achieves the best performance across all datasets in both settings. Notably, although all models experience performance drops when user interaction data is limited, the decline in our method is significantly smaller, especially on complex datasets such as AM-Luxury and AM-Scientific. For example, on AM-Luxury, our method only drops 0.0022 in Recall@20, whereas BERT4Rec drops 0.0341 and RecMind drops 0.0380. This indicates that our model can better capture user intent with limited supervision. We attribute this advantage to the dual-view representation and the semantic richness preserved in our LLM-based design, which allows our model to generalize better under cold-start conditions. These results further confirm the effectiveness of our approach in improving recommendation robustness in real-world scenarios.

\subsubsection{Scalability Analysis}

To better understand the practical feasibility of our proposed model, we conduct additional experiments to evaluate its training and inference efficiency under different data scales. We select three Amazon datasets of varying sizes---AM-Luxury (small), AM-Toys (medium), and AM-Food (large)---to simulate realistic scenarios with increasing user-item interaction volumes.

We measure and report the following metrics:
\begin{itemize}
    \item Training time per epoch (in seconds)
    \item Validation inference time (in seconds)
    \item Peak GPU memory usage (in MB)
\end{itemize}

The results are summarized in \textbf{Table~\ref{tab:scalability}}. The training and inference times grow approximately linearly with the dataset size, while the peak memory consumption remains relatively stable across different datasets. These findings demonstrate that our model is computationally efficient and scalable to larger recommendation tasks.

We also observe that the inference time is longer than the training time per epoch. This is because, during inference, the model must compute scores for all candidate items for each user and perform a complete ranking. In contrast, training only requires calculating the loss based on sampled interactions. The full ranking process increases the computational workload and inference latency.

\begin{table}[h]
    \centering
    \caption{Training and Inference Efficiency on Datasets of Different Sizes.}
    \label{tab:scalability}
    \resizebox{0.8\textwidth}{!}{
    \begin{tabular}{lccc}
        \toprule
        Dataset & Training Time (s/epoch) & Inference Time (s) & Peak Memory (MB) \\
        \midrule
        AM-Luxury & 4.46 & 5.02 & 5002.08 \\
        AM-Toys & 38.14 & 51.13 & 20018.38 \\
        AM-Food & 233.85 & 327.18 & 26096.70 \\
        \bottomrule
    \end{tabular}
    }
\end{table}

Overall, these experimental results verify that the proposed graph-attentive LLM recommender maintains good computational efficiency and can be effectively deployed in large-scale recommendation systems.

\subsubsection{Case Study: Graph-Aware Recommendation Reasoning}

We present a case study based on a single user from the AM-Toys dataset to demonstrate how our model leverages structural connectivity between users and items during recommendation.

\begin{figure}[h]
    \centering
    \includegraphics[width=0.8\linewidth]{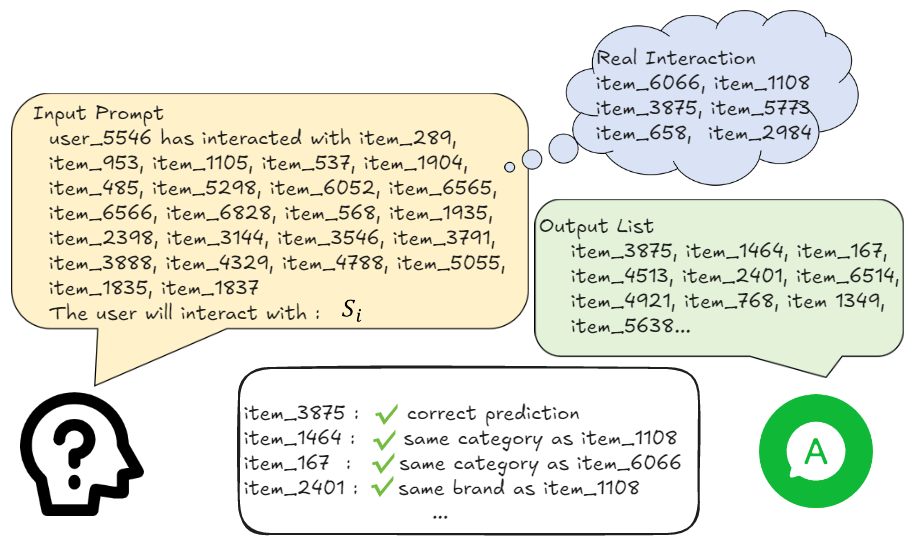}
    \caption{A case study showing the model's recommendation results and their connections to the user's historical interactions.}
    \label{fig:casestudy}
\end{figure}

As illustrated in Figure~\ref{fig:casestudy}, the model receives a prompt indicating that \texttt{user\_5546} has previously interacted with four items: \texttt{item\_1105}, \texttt{item\_537}, \texttt{item\_1904}, and \texttt{item\_485}. Based on this interaction history, the model is prompted to generate a personalized recommendation list.

The predicted list includes items such as \texttt{item\_3875}, \texttt{item\_2464}, and \texttt{item\_2401}. Notably, \texttt{item\_3875} appears in the user's actual future interactions, confirming the correctness of the prediction. Even for items not directly interacted with, we observe meaningful structural or semantic connections. For instance, \texttt{item\_1464} shares the same category as \texttt{item\_1108}, and \texttt{item\_2401} belongs to the same brand as \texttt{item\_1108}.
This case highlights how the model captures both direct and indirect connections in the interaction graph to generate meaningful and interpretable recommendations.

% \vspace{-0.2cm}
\section{Related Work}\label{related_work}

Recommender systems (RS) play a crucial role in various applications, helping users navigate vast amounts of information efficiently. Traditional RS methods have been extensively studied and widely applied, laying a strong foundation for the field. 
Recently, the emergence of LLMs has introduced new possibilities, broadly categorized into two approaches: (1) leveraging LLMs for deep representation learning to enhance user and item embeddings and (2) directly applying generative LLMs to construct recommendation logic, enabling more adaptive and context-aware recommendations.

\subsection{Traditional Recommender System}

Traditional recommendation algorithms began with similarity-based methods, particularly Collaborative Filtering (CF), which exploits user-item interactions~\cite{Koren2009MatrixFT, Salakhutdinov2008BayesianPM, Rendle2009BPRBP}. Early CF models, such as Matrix Factorization (MF)~\cite{Koren2009MatrixFT}, effectively captured latent preferences but struggled with data sparsity and linearity constraints.
The rise of neural networks introduced models like Neural Collaborative Filtering~\cite{NCF}, which replaced inner product-based interactions with multi-layer perceptrons for greater expressiveness.
More recently, Graph Neural Networks (GNNs) have become prominent in recommendation~
\cite{NGCF,he2020lightgcn,URAC}. Neural Graph Collaborative Filtering~\cite{NGCF} and LightGCN~\cite{he2020lightgcn} propagate user-item interaction signals through graph structures, effectively capturing high-order relationships.
Beyond model architecture, research has expanded into diversified open problems, such as multi-modal recommendation, leveraging textual, visual, and knowledge graph data~\cite{he2016vbpr,he2024boosting}; denoising-based recommendation, addressing noisy interactions~\cite{denoising}; debias recommendation, tackling issues like popularity bias~\cite{cmm1,cmm2}; cold-start recommendation, recommending to new user without historical interactions~\cite{gorec,MILK}.
Another direction is feature engineering, which enhances recommendation performance by refining input representations~\cite{lian2018xdeepfm,lin2022adafs}. Feature selection techniques identify the most informative features to reduce noise and improve generalization~\cite{ying2024feature,wang2024knockoff,gong2025neuro,ying2025survey,ying2024revolutionizing}, while feature transformation, such as autoencoders and reinforcement learning-based approaches, enable the construction of more expressive feature spaces~\cite{ying2024unsupervised,ying2023self,gong2025evolutionary,wang2025towards,ying2024topology,gong2025unsupervised,gong2025sculpting,hu2024reinforcement,bai2025privacy，wang2025llm}. These methods contribute to the robustness and adaptability of recommendation systems across various domains.
This evolution from CF to neural and graph-based models, alongside advances in feature engineering and diverse research topics, reflects the ongoing advancement in recommender systems.
% This evolution from CF to neural and graph-based models, alongside diverse research topics, reflects the ongoing advancement in recommender systems.

% CF-based methods are widely used in recommendation \cite{Koren2009MatrixFT, Salakhutdinov2008BayesianPM, Rendle2009BPRBP, Chen2020RevisitingGB, Wu2020LearningTT}.
% Rich historical interaction records are key to the success of these methods.
% However, CF-based methods encounter a significant hurdle with the cold-start problem, where the model struggles to provide effective recommendations for users or items with insufficient historical interaction data. 
% This issue can be classified into two types: completely cold-start and incompletely cold-start, depending on whether there are any previous interaction records \cite{wei2017collaborative}. 

\subsection{Discriminative LLMs for Recommender System}

In deep representation, discriminative language models like BERT are widely used for fine-tuning and pre-training, integrating specific domain data features to enhance the performance of recommender systems.
For instance, U-BERT~\cite{qiu2021u} leverages content-rich domain data to learn user representations, compensating for the scarcity of behavioral data. Similarly, UserBERT~\cite{wu2021empowering} includes two self-supervised tasks for pretraining on unlabeled behavior data. Additionally, BECR~\cite{yang2022lightweight} combines deep contextual token interactions with traditional lexical word matching features.
Notably, the "pretrain-finetune" mechanism plays a crucial role in sequence or session-based recommender systems, like BERT4Rec~\cite{sun2019bert4rec} and RESETBERT4Rec~\cite{zhao2022resetbert4rec}. UniSRec~\cite{hou2022towards} develops a BERT fine-tuning framework that links item description texts.
In content-based recommendations, especially in the news domain, models like NRMS~\cite{wu2021empowering}, Tiny-NewsRec~\cite{yu2021tiny}, and PREC~\cite{liu2022boosting} enhance news recommendations by leveraging LLMs, addressing domain transfer issues, or reducing transfer costs. Research by Penha and Hauff~\cite{penha2020does} shows that BERT, even without fine-tuning, effectively prioritizes relevant items in ranking processes, illustrating the potential of large language models in natural language understanding.
DWSRec~\cite{zhang2024dual} relies only on pre-trained text embeddings without the need for ID embeddings. By applying dual-view whitening, it enhances the isotropy of embeddings while preserving semantic information.
LLMRec~\cite{wei2024llmrec} enhances recommender systems by leveraging LLMs for graph-based augmentation and denoised data robustification, effectively mitigating data sparsity.
CoLLM~\cite{zhang2025collm} integrates collaborative filtering information as a separate modality into large language models (LLMs) via an external mapping module, aligning collaborative embeddings with text inputs.

\subsection{Generative LLMs for Recommender System}

Recent advances in generative models have combined neural generation with symbolic reasoning, enabling more interpretable and structured decision-making~\cite{ying2025bridging,gong2025agentic}. Building on this, generative LLMs have shown strong potential in recommender systems through prompting, fine-tuning, and routing~\cite{wang2025llm}.
%By transforming recommendation tasks into natural language processing tasks, researchers have successfully utilized these models' inherent rich knowledge and strong representational capabilities, providing new perspectives and methods for handling complex recommendation tasks. Utilizing their advantages in language understanding and generation, these models offer more precise and personalized recommendations. 
Notable works and advancements include:
Liu et al.~\cite{liu2023chatgpt} conducted a comprehensive assessment of ChatGPT's performance in five key recommendation tasks. 
%They proposed a universal recommendation prompt construction framework, to enhance the understandability and assessability of the recommendation results.
Sanner et al.~\cite{sanner2023large} designed three different prompt templates to evaluate the enhancement effect of prompts, finding that zero-shot and few-shot strategies are particularly effective in preference-based recommendations using language.
Sileo et al.~\cite{sileo2022zero} and Hou et al.~\cite{hou2023large} focused on designing effective prompt methods for specific recommendation tasks.
%such as movie recommendations and sequential recommendations.
Gao and team~\cite{gao2023chat} developed ChatREC around ChatGPT, an interactive recommendation framework that understands user needs through multiple rounds of dialogue.
Petrov and Macdonald~\cite{petrov2023generative} introduced GPTRec, a generative sequence recommendation model based on GPT-2.
%, which shows higher flexibility and efficiency compared to discriminative models like BERT4Rec.
Kang and colleagues~\cite{kang2023llms} explored formatting user historical interactions as prompts and assessed the performance of LLMs of different scales.
PageLLM~\cite{wang2025enhanced} introduces a multi-grained reward model to fine-tune the LLM using reinforcement learning from human feedback.
Dai et al.~\cite{dai2023uncovering} designed templates for various recommendation tasks using demonstration example templates.
%, demonstrating that context learning methods significantly enhance the recommendation effectiveness of LLMs.
Bao et al.~\cite{bao2023tallrec} developed TALLRec, which demonstrates the potential of LLMs in recommendation domains through two-stage fine-tuning training.
Ji et al.~\cite{ji2023genrec} presented GenRec, a method that leverages the generative capabilities of LLMs to directly generate the target of recommendations.
In specific scenarios like online recruitment, generative recommendation models such as GIRL~\cite{zheng2023generative} and reclm~\cite{friedman2023leveraging} demonstrated enhanced explainability and appropriateness in recommendations.
Li et al. ~\cite{li2023pbnr} described user behaviors and designed prompts in news with PBNR.
%, showing the efficacy of fine-tuned LLMs in content recommendations.
Wang et al.~\cite{wang2022towards} proposed UniCRS, a design based on knowledge-enhanced rapid learning.
%, providing a new direction for unified session recommendation systems.
HSTU~\cite{zhai2024actions} is a generative recommendation model that treats recommendation as a sequence generation task, achieving high accuracy and efficiency on large-scale recommendation data.
RecMind~\cite{wang2023recmind} is an LLM-based recommendation agent that makes zero-shot personalized recommendations by using external tools and a self-inspiring planning method to better use past information.

%The application of large language models in recommendation systems not only expands the scope of technological application but also provides new perspectives and methods for optimizing and innovating recommendation algorithms. Moreover, the diverse use and in-depth research of large language models in this field are rapidly evolving, enhancing the performance of recommendation systems and offering new directions and inspirations for future research and applications.

% \vspace{-0.2cm}
\section{Conclusion}
% \vspace{-0.1cm}
We tackle a key issue in recommender systems: how to integrate LLM and graph structures into recommendations. To this end, we propose a graph attentive LLM generative recommender system. By introducing new prompting methods and graph-structured attention mechanisms, we can effectively integrate the complex relationships and background information between users and items into the model. We first designed a natural language prompt that can reflect the relationship between users and items and embed the 2-order relationship between items into it. Next, we improved the attention mechanism of LLM to model complex graph structure information. Through experiments, we validate the effectiveness of our method. The experimental results show that our model has significantly improved recommendation accuracy and personalization compared to traditional recommender systems.
Considering these innovations, our approach provides a new technological path for developing more efficient and intelligent recommender systems. 
Meanwhile, these methods demonstrate new perspectives and ideas in applying LLM to recommender systems and a wider range of fields. This promotes the development of recommender systems and provides strong support and inspiration for using LLM in various complex application scenarios.
\bibliographystyle{ACM-Reference-Format}
\bibliography{sample-base}

\end{document}